\documentclass[letterpaper]{article} 
\usepackage[preprint]{aaai2027}  
\usepackage[hyphens]{url}  
\usepackage{graphicx} 
\usepackage{natbib}  
\usepackage{caption} 
\usepackage{amsmath}
\usepackage{amssymb}
\usepackage{colortbl}
\usepackage{booktabs}
\usepackage{multirow}
\usepackage{tabularx}
\usepackage{siunitx}
\usepackage{tcolorbox}
\usepackage{pifont}  
\usepackage[table]{xcolor}
\tcbuselibrary{breakable}
\definecolor{hcgray}{gray}{0.90}
\newcommand{\hcrow}{\rowcolor{hcgray}}
\newcommand{\rlab}[1]{\rotatebox[origin=c]{90}{\textit{#1}}}
\newcommand{\abl}[1]{\quad\textit{#1}}
\newcommand{\cmark}{\textcolor{green!60!black}{\ding{51}}}   
\newcommand{\xmark}{\textcolor{red!70!black}{\ding{55}}}     
\newcommand{\pmark}{\textcolor{green!45!black}{(\ding{51})}} 

\usepackage{xcolor}
\usepackage{soul}
\usepackage{float}
\definecolor{fixyellow}{rgb}{1,0.96,0.6}   

\tcbuselibrary{breakable,skins}
\tcbset{smbprompt/.style={
  breakable, enhanced jigsaw,
  colback=gray!5, colframe=black!55, boxrule=0.4pt, arc=1pt,
  left=5pt, right=5pt, top=4pt, bottom=4pt,
  fonttitle=\bfseries, coltitle=black, colbacktitle=gray!20,
}}
\title{Can Agent Memory Systems Track Evolving State?}

\author{
    Xinyi Fan\equalcontrib,
    Miri Liu\equalcontrib,
    Ruozhen Yang,
    Siru Ouyang,
    Jiawei Han
}
\affiliations{
    University of Illinois Urbana-Champaign\\
    \{xfan31, miri3, ruozhen2, siruo2, hanj\}@illinois.edu
}

\begin{document}
\maketitle

\begin{abstract}

As LLM-based agents are deployed for longer and higher-stakes tasks, their memory systems continue to have crucial gaps. While existing memory benchmarks focus largely on recall-shaped tasks, we argue an effective memory system must track the evolving state of the world; as facts, constraints, and decisions are revised over a long interaction, answers must reflect the current state and not a superseded one. We define this capability as state tracking and instantiate it in \textsc{StateMemBench}, a benchmark of 234 multi-session scenarios spanning two conversation-length regimes. Its closed-pool grading scores whether an answer reflects the current state, the superseded state, or fails otherwise, separating state-tracking failures from other errors by construction. Our analysis shows that this task is challenging for existing memory systems, retrieval-augmented baselines, and long-context baselines. We then present \textsc{StateMem}, a state-first memory method that explicitly tracks supersession and relational dependencies, and show it improves current-state accuracy over the strongest same-backbone baseline by 1.8$\times$ (0.205 $\rightarrow$ 0.363) on DeepSeek-V4-Flash and over the strongest memory system by 1.6$\times$ (0.149 $\rightarrow$ 0.233) on Qwen-3.5-9B, while remaining competitive with the long-context baselines. Finally, we show the same state approach can be applied as a lightweight single-call wrapper over existing memory systems, lifting current-state accuracy by +32 to +67 points on \textsc{StateMemBench} across six memory and retrieval backends. A length- and cost-matched control attributes +15 to +32 of those points to state structure rather than added context.

\end{abstract}

\section{Introduction}
\label{sec:introduction}
\begin{figure}
    \centering
    \includegraphics[width=0.76\linewidth]{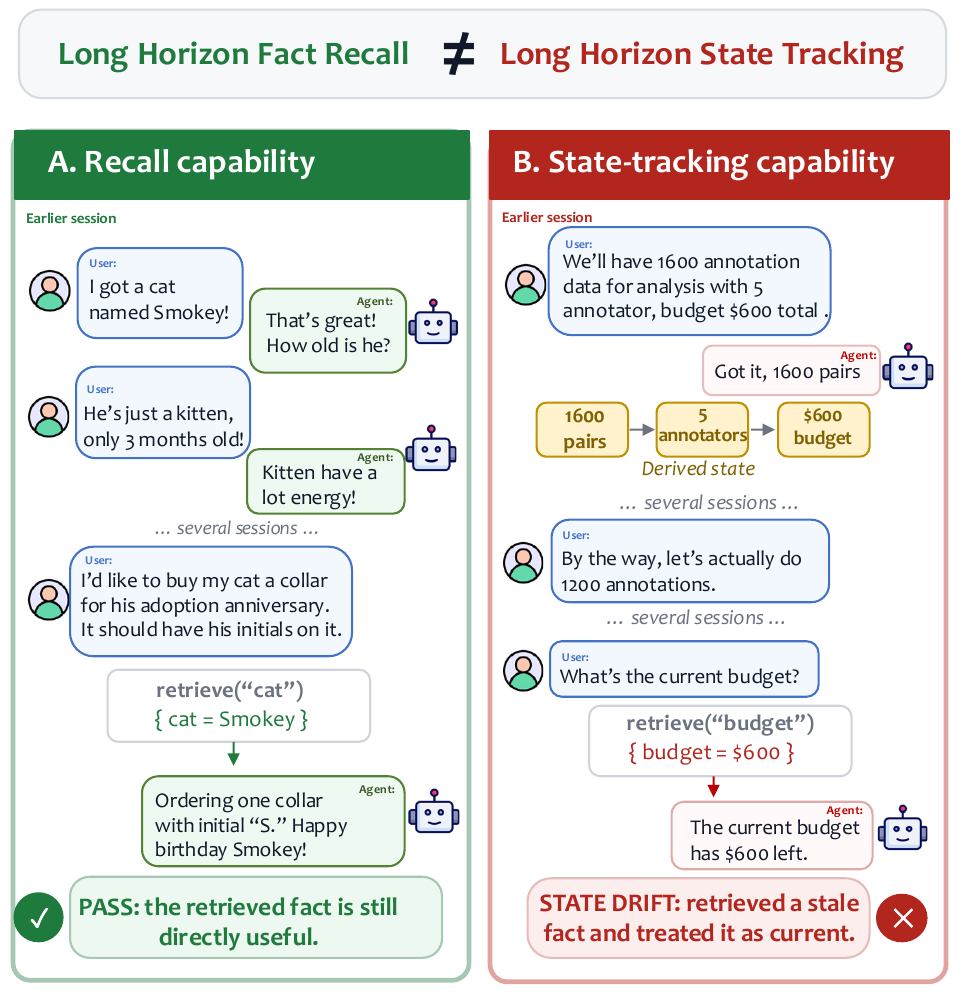}
\caption{\textsc{StateMemBench} targets state tracking: maintaining currently operative values of facts, rules, and derived quantities under cross-session revision, separable from recall.}
    \label{fig:teaser}
\end{figure}

Agents are now capable of (and deployed for) tasks that take place over multiple sessions, from managing end-to-end research pipelines \citep{gottweis2025towards, lu2024ai} to autonomously working on software codebases \citep{wu2024devin, yang2024swe}. The ability of an agent to keep working is, firstly, a crucial benchmark of its capabilities \citep{kwa2025measuring}, and, secondly, something that requires memory. Accordingly, many different strategies for managing agent memory have emerged, from extending the model's effective context \citep{ding2024longrope} to external memory systems \citep{memgpt, mem0}, just as many benchmarks have been proposed to evaluate these systems \citep{wu2024longmemeval, jiang2025personamem, locomo}.

These systems and benchmarks focus heavily on optimizing recall of relevant facts, which is undoubtedly an important component of memory for humans and LLMs alike. However, we argue that as what we ask of agents becomes more time-consuming and complex, the ability of an agent to \textbf{track state} as it evolves is a crucial capability, and one that current systems largely overlook. For instance, as shown in Figure~\ref{fig:teaser}, we want agent memory systems to be able not only to recognize that the up-to-date number of annotations is 1200, but also to be able to recognize the state-level implications of that change.

We call this failure \textbf{state drift}: the relevant fact is present in the assembled context, but the agent acts on a stale or incomplete version of it. The obstacle is not \emph{retrieving} the fact, but rather \emph{tracking} its relevance to current state, which we formalize in \S\ref{sec:problem}. While some recent benchmarks and memory systems \citep{wu2026longmemeval, chao2026stale} have begun to center state as a first-class memory capability, and almost all systems update or revise memory in some sense, none cleanly isolates state tracking from the various other errors with which it co-occurs. We close this gap, and make four important contributions to agent memory:

\begin{itemize}
    \item We characterize \textbf{state tracking} as a capability distinct from recall, and give a labeling procedure that isolates state drift from other errors. Drift leads the failure distribution on several established memory benchmarks and persists even under perfect retrieval (\S\ref{sec:problem}; rates are judge-labeled upper bounds).
    \item We propose \textsc{\textbf{StateMemBench}}, a multi-domain benchmark targeted specifically at the problem of state tracking, with 234 multi-session scenarios spanning shorter and longer conversation lengths across three domains.
    \item We propose \textsc{\textbf{StateMem}}, a straightforward, state-first method, and show it improves over the strongest same-backbone baseline by 1.8$\times$ on DeepSeek (0.205 $\rightarrow$ 0.363) and over the strongest memory system by 1.6$\times$ on Qwen-9B (0.149 $\rightarrow$ 0.233), beating same-backbone long-context on both (0.149 on each).
    \item We show the state-tracking axis is \textbf{separable}: a lightweight answer-time wrapper form of StateMem improves every backend it is applied to, with a length- and cost-matched control isolating the share attributable to state structure rather than added context (\S\ref{sec:wrapper}).
\end{itemize}

\section{Related Work}
\label{sec:related}
\paragraph{Long-term memory systems for LLM agents.}
Many existing memory systems optimize recall, including through virtual memory hierarchies with explicit read/write operations \citep{memgpt}, reflection-based memory streams \citep{park2023generative}, linked memory structures \citep{xu2025amemagenticmemoryllm}, or periodic summarization \citep{wu2025resum}. However, we consider recall to be orthogonal to state tracking. Even under perfect retrieval, a system still has to resolve which of the surfaced facts is currently valid (\S\ref{sec:perfect-retrieval}). Fewer systems attempt to address this. Zep \citep{zep} attaches valid-time intervals to knowledge-graph edges and invalidates an edge on detecting a contradiction, which allows the system to track the validity of a single fact. The concurrent STALE \citep{chao2026stale} also targets state tracking and propagates updates, but does so through an LLM adjudicator over a predefined schema, whereas \textsc{StateMem} uses a deterministic, LLM-free pass over a dependency graph. Additionally, STALE constructs conflicts \emph{implicitly}, so detecting them requires commonsense inference, while our evaluation testbed \textsc{StateMemBench} supplies conflicts \emph{explicitly}, isolating state maintenance from that inference.

\paragraph{Dialogue state tracking.}
The term \emph{state} has been extensively used in the task of \textbf{dialogue state tracking} (DST). Our work differs from this task in what \emph{state} is and what is evaluated. DST prescribes the representation, whether as slot--value pairs over a domain ontology \citep{williams2013dialog, budzianowski2018multiwoz}, or its schema-guided \citep{rastogi2020towards} and open-vocabulary \citep{heck2020trippy} relaxations, and evaluates that \textit{representation} directly. On the other hand we do not require state to take any particular form; instead, state is what a memory system must maintain to answer correctly, and evaluation is purely behavioral. Additionally, DST corpora are cooperative, accumulating a user goal monotonically within one dialogue, and \textsc{StateMemBench} adversarially revises facts and user or agent decisions across multiple sessions to induce and measure stale-state failures \citep{kim2022mismatch, bae2022keep}.

\paragraph{Memory benchmarks for LLM agent memory.}
There are many memory benchmarks for LLM agent memory. Long conversational memory benchmarks \citep{locomo, wu2024longmemeval, li2026locomo} evaluate how well a model can answer questions about the user's messages, from surface-level factual recall to more complicated implicit intent. Some focus on personalizing responses to the user \citep{jiang2025personamem}, which overlaps with long conversational memory more generally. Similarly, user interaction benchmarks \citep{yao2024tau, barres2025tau} evaluate tool-calling agents on their customer service skills, providing a controlled environment for measuring constraint adherence, but these are single-session, so agents do not necessarily need memory. Moving further from user dialogue, recent benchmarks \citep{zhao2026ama, he2026memoryarena} evaluate how well an agent can use its memory system to accomplish tasks autonomously or semi-autonomously; notably, \citet{wu2026longmemeval} evaluate whether agentic memory systems can learn through interaction how the environment changes, which they call dynamic state tracking. This targets learning the \emph{environment's} dynamics as a world model, measured over trajectories of up to $115$M tokens, thus entangling the task with large-scale retrieval. \textsc{StateMemBench} aims to isolate whether a memory system that \emph{already holds} the relevant facts maintains the currently operative state, separately from retrieval and reasoning.

\section{Problem Formulation: State Drift}
\label{sec:problem}

Long-horizon agents fail even when the relevant information is in context. Precisely, \textbf{state drift} occurs when a fact, constraint, or dependency established or updated across prior context is present in the assembled context, but the agent acts on a stale or incomplete version of the state rather than the one in force at decision time. This separates
failing to \textbf{maintain} state from failing to \textbf{acquire} it, as better retrieval does not fix the former.

\subsection{Definition and Labeling}

We consider drift in the context of possible neighboring failures. For instance, in a retrieval failure, the gold fact is not present, and in a schema mismatch, the answer is correct but badly formatted. For harder neighbors like reasoning failure, we consider tests like the following: the trace must demonstrably engage the current, up-to-date value, then err. Each $(case, turn, slot)$ failure is assigned to drift only once all other readings are excluded; unassignable points are dropped, not defaulted (\emph{confirmed failures} = this filtered set). An adversarial filter plus cross-model-family judging (Appendix~\ref{app:drift-label-validation}) bias every exclusion \emph{against} drift. We apply it on LongMemEval oracle \citep{wu2024longmemeval}, MemoryArena (MA) \citep{he2026memoryarena}, and $\tau^2$ \citep{barres2025tau}.


Two judge passes agree on the binary drift decision at
$\kappa{=}0.67$ on LongMemEval and $91.6\%$ raw on MA-shopping
(prevalence-deflated $\kappa$; PABAK${=}0.83$); a cross-family judge
(GPT-4o) agrees at $\kappa{=}0.37$ (Table~\ref{tab:agreement}). Two human annotators,
labeling under a structured protocol and, separately, describing
failures blind to the taxonomy, recover the same categories, every blind description mapping to an existing bucket. Where humans depart from the judge they assign \emph{less} drift, so judge rates are a ceiling within an already-conservative procedure (full agreement statistics in Appendix~\ref{app:drift-label-validation}).

Drift is also not a reasoning shortfall. On $n{=}50$ paired LongMemEval items, enabling DeepSeek-V4-Flash's reasoning trace does not help (84.0\%$\to$76.0\%; McNemar $p{=}0.34$; Appendix~\ref{app:drift-label-validation}), while explicit state maintenance on the same task does (\S\ref{sec:wrapper}).

\subsection{Drift Persists Under Perfect Retrieval}
\label{sec:perfect-retrieval}
On LongMemEval oracle, recall is $1.0$ by construction, so no residual
  failure can be retrieval in disguise. DeepSeek-V4-Flash fails $36$ of
  $306$ questions;\footnote{The oracle pass covers LongMemEval's three cross-session
  question types (multi-session, temporal-reasoning, and
  knowledge-update) since single-session questions
  have no state to lose. The memory-system evaluation of
  \S\ref{sec:results-lme} uses all $500$ questions.} judges from two model families both confirm $44.4\%$
  ($16/36$; Wilson 95\% CI $[30\%, 60\%]$) as state drift --- the largest confirmed failure category
  under guaranteed evidence (Table~\ref{tab:cross-bench-failure}). Drift notably concentrates where decisions integrate multiple sources; failures
  of multi-session questions drift at nearly three times the rate of
  knowledge-update failures ($71.4\%$ vs.\ $25.0\%$; $10/14$ vs.\ $2/8$,
  identical under either judge family). For details, see
  Appendix~\ref{app:drift-case}.
 
\subsection{Cross-Benchmark Prevalence}
\label{sec:prevalence}
 

Drift is substantial but not universal across benchmarks
(Table~\ref{tab:cross-bench-failure}): it leads on MA-shopping ($63.5\%$) and on $\tau^2$-bench-Z ($10/16$ failures), but falls to $19.0\%$ on MA-travel, where comprehension failure dominates ($46.5\%$). Note that rates are LLM-judge labels and that human annotators tend to assign
less drift (see Appendix~\ref{app:drift-label-validation}). However, this
labeling cannot tell us why the wrong value wins. Judges
can agree on whether a failure is drift, but not on its mechanism (such as a
louder competitor or a genuine revision left unapplied) with mode
$\kappa$ near zero. To study the mechanism we must control it more explicitly. 
We therefore build \textsc{StateMemBench}, where each scenario's failure mechanism is fixed by construction and the superseded value is a scored outcome (Table~\ref{tab:benchmark-comparison} compares against
existing benchmarks).

\begin{table}[t]
\centering
\definecolor{hcgray}{gray}{0.90}
\small
\caption{\textbf{Cross-benchmark failure-mode distribution.} Each
cell is the share of that benchmark's confirmed failures assigned to
each bucket; the drift column is
highlighted. $^\dagger$On LongMemEval
oracle every gold fact is in the prompt by construction, so no failure
there can be a retrieval failure. The small-$n$ $\tau^2$-bench-Z result
($n{=}16$) is reported in \S\ref{sec:prevalence}. Labeling, adjudication,
and per-row details: Appendix~\ref{app:drift-label-validation}.}
\label{tab:cross-bench-failure}
\setlength{\tabcolsep}{3pt}
\renewcommand{\arraystretch}{0.95}
\resizebox{\columnwidth}{!}{%
\begin{tabular}{lr>{\columncolor{hcgray}}rrrrrr}
\toprule
Benchmark & $n$ & \textbf{drift} & retr. & comp. & schema & reason & FA \\
\midrule
\multicolumn{8}{l}{\emph{MemoryArena}}\\
Shop.   & 200 & \textbf{63.5} & 22.5 &  2.5 &  0.0 &  9.5 & -- \\
Search  & 200 & \textbf{38.5} & 25.5 & 11.0 & 17.5 &  4.5 & -- \\
Travel  & 200 & \textbf{19.0} & 11.0 & 46.5 & 21.0 &  0.0 & -- \\
\midrule
\multicolumn{8}{l}{\emph{LongMemEval oracle}}\\
LME$^\dagger$ & 36 & \textbf{44.4} & 0.0 & 0.0 & 25.0 & 2.8 & 27.8 \\
\bottomrule
\end{tabular}%
}
\end{table}

\section{\textsc{StateMemBench}}

\begin{table*}[!t]
\centering
\small
\setlength{\tabcolsep}{4.5pt}
\renewcommand{\arraystretch}{1.15}
\caption{Comparison with agent-memory and long-horizon benchmarks with state-tracking evaluation. \pmark~=~partial satisfied. \emph{Scale} is reported in each
benchmark's native unit. Column
definitions and per-benchmark justifications are in
Appendix~\ref{app:benchmark-comparison}.}
\label{tab:benchmark-comparison}
\begin{tabular}{l cc cccccc}
\toprule
\textbf{Benchmark} & \shortstack{Multi-sess.\\dialogue} & \shortstack{Scale\\(native unit)} & \shortstack{Updates\\central} & \shortstack{Super-\\session} & \shortstack{Verifiable\\gold} & \shortstack{Drift\\scored} & \shortstack{Anti-update\\controls} & \shortstack{Paired\\horizons} \\
\midrule
LoCoMo \shortcite{locomo}                     & \cmark & $\sim$300 turns          & \xmark & \xmark & \xmark & \xmark & \xmark & \xmark \\
LongMemEval \shortcite{wu2024longmemeval}     & \cmark & $\sim$40--80 sess.\ ($S$) & \pmark & \pmark & \xmark & \xmark & \xmark & \xmark \\
MemoryAgentBench \shortcite{hu2025evaluating} & \pmark & varies                   & \pmark & \pmark & \pmark & \xmark & \xmark & \xmark \\
MemoryArena \shortcite{he2026memoryarena}     & \pmark & varies                   & \xmark & \xmark & \cmark & \xmark & \xmark & \xmark \\
STATE-Bench \shortcite{statebench2026}        & \xmark & 450 tasks                & \xmark & \xmark & \pmark & \xmark & \pmark & \xmark \\
\rowcolor{hcgray}
\textbf{StateMemBench (ours)}                 & \cmark & $\sim$600 turns          & \cmark & \cmark & \cmark & \cmark & \cmark & \cmark \\
\bottomrule
\end{tabular}
\end{table*}
\label{sec:benchmark}
We release \textsc{StateMemBench}, a benchmark that aims to directly measure memory systems' ability to \textbf{track state.} We explain how failure modes are defined and computed, and we give an overview of the benchmark domains, the scenario generation pipeline, and our validation gates.

\subsection{Failure Modes as Policy Divergence}
\label{sec:taxonomy}
We define failure modes mechanically rather than by hand-defining a taxonomy. We generate each scenario as a symbolic event program, or a sequence of typed state operations (rule declarations, value updates, scoped exceptions, commitments, and retractions) over ground, derived, and declared state. The correct answer to a probe is computed by replaying the program through a deterministic evaluator. Alongside the evaluator, we implement a family of lazy reader policies. These are small executable heuristics which each answer the probe according to some possible failure of a memory system. For instance, systems may trust the most recent mention, trust the most frequent value, return a stated value without recomputing it, or discard an anchored decision too eagerly. We get our trap scenarios exactly when one or more policies disagree with full replay. Those disagreeing policies then form the scenario's failure-mode signature. They instantiate five failure modes:

\paragraph{\textbf{Status}} errors occur when a value changes (a tiered rule may move tiers, or a scoped override may lapse), but a reader remains anchored on the earlier, louder commitment and thus answers with the superseded value.

\paragraph{\textbf{Salience}} errors occur when the valid value is present but a more frequently or more prominently mentioned competitor wins. Unlike a status error, nothing was actually superseded.

\paragraph{\textbf{Sequence}} errors occur when a stated derived value is not recomputed after one of its inputs changes. The reader returns the stale derivation instead of carrying the update forward through the stated dependency.

\paragraph{\textbf{Compound}} errors arise when several mechanisms interact.

\paragraph{\textbf{Anti-trap}} scenarios test the opposite direction. An anchored answer will remain correct, and only an over-eager invalidation policy will result in divergence.

\subsection{Domains}
\label{sec:state-mem-bench_datasets}
We instantiate the benchmark in three domains (research, shopping, and personal finance), chosen for three reasons: (1) shopping and personal finance capture single user--assistant state, while research captures collaborative project state; (2) each surfaces a unique mix of categorical choices and constraints; and (3) all three can be grounded in readily available public data. Note that public data supplies only the surface of a scenario, while state values are independently sampled, and no probe is answerable from the source material. For more information, see Appendix~\S\ref{app:domains}.

\subsection{Scenario Generation}
\label{sec:scenario_generation}
We design each scenario so causal dependencies are stated explicitly (i.e., "Since we have 1600 annotators, we'll need 5 external annotators"), so a system is not required to infer that one fact bears on another. This isolates state tracking from the need to discover or assume relationships, letting us focus on whether a memory system actually maintains state as context evolves. The released benchmark has two length conditions: \textbf{Set~A} (short) contains 190 single-probe scenarios of 18 sessions each (median 165 turns, ${\sim}3$k tokens), and \textbf{Set~B} (long) contains 44 scenarios, each fusing three trap threads of different modes into one $\sim$38-session conversation (median 599 turns, $3.6\times$ Set~A in turns; ${\sim}7$--$15$k tokens) probed by a three-part question --- \textbf{234 scenarios, 322 graded probes} in total. In Set~B, each thread's gold-bearing events are buried among the \emph{other} threads' live state, making Set B significantly more challenging. We detail scenario validation in Appendix~\S\ref{app:smb-validation}.

\paragraph{Grounding and Rendering}
Each sampled program is grounded with surface material from the domain's real data (\S\ref{sec:state-mem-bench_datasets}) and a strong LLM (\texttt{sonnet-4.6}) renders it into natural multi-session dialogue. The grounding supplies only surface vocabulary and never decides trap semantics, so it cannot contaminate a probe. We then programmatically verify that every load-bearing fact appears in its assigned session and that no banned phrase leaks, re-rendering any scenario that fails. Full pipeline details are in Appendix~\ref{app:scenario-gen}.

\paragraph{Probe Design}
All scenarios carry a probe (the final turn, meant to test state tracking) and a gold truth (the correct response based on state changes). Probes are closed-pool, meaning that at scenario generation time we construct a set of \textit{plausible} answers, though these are not shown at probe time. In addition to the gold truth, the pool contains the targeted policy's computed \textbf{drift answer}, which represents the value a reader that does not track state would anchor on, as well as relevant neutral distractors (3--4 options total). This means we can separate badly-tracked answers (\textit{drift}) from totally-incorrect ones (\textit{other}). The drift labels enable the error analysis in \S\ref{sec:error} over the entire benchmark.

\section{\textsc{StateMem}}
\label{sec:method}

\begin{figure}[h]
    \centering
    \includegraphics[width=\linewidth]{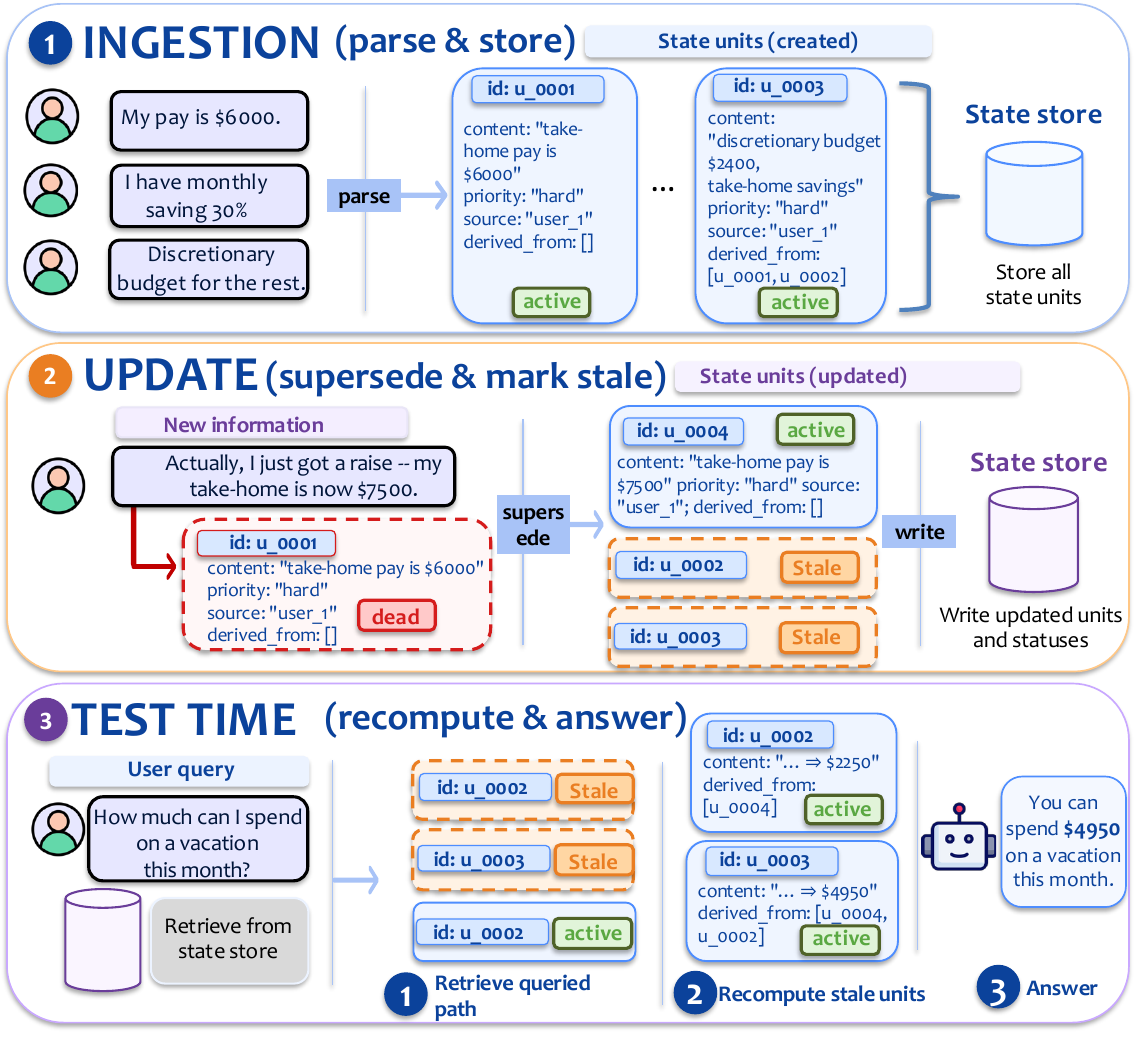}
    \caption{\textsc{StateMem} has three stages: an ingestion stage in which a conversation turn is parsed into state units, an update stage that updates the value of existing units, and a test time stage where the agent draws upon the valid state to answer the question. Here, we show how \textsc{StateMem} would handle a finance-based scenario.}
    \label{fig:mechanism}
\end{figure}

We introduce \textsc{StateMem}, a straightforward, state-first approach to memory that represents multi-turn conversational memory as a collection of
structured \emph{state units}. As shown in Figure~\ref{fig:mechanism}, StateMem operates in three stages: \textbf{ingestion} parses a turn into
state units, \textbf{update} revises existing units, and \textbf{test time}
uses the resulting coherent state to answer a question. Full prompts are in
Appendix~\ref{app:prompts}.

\subsection{Ingestion}
\label{sec:ingestion}
A \texttt{TurnEncoder} applies one LLM call per conversational turn $t$,
parsing it into zero or more state units, with the current compact rendering
of all active units supplied as context. State stays compact at our scale,
but the step is modular and can be swapped for slice-based retrieval of the
\texttt{StateStore} as state grows. The encoder is domain-agnostic, admitting
an optional \texttt{task\_context}
(Appendix~\ref{app:turn-encoder-prompt}), though we report all results
without it.
Each state unit is a tuple
$ u = (\textit{id},\ \textit{content},\ \textit{priority},\ \textit{source},\ \textit{deps}), $
with \textbf{id} a unique identifier, \textbf{content} a free-form field,
$\textbf{priority} \in \{\text{hard}, \text{soft}\}$, \textbf{source}
recording the originating turn and speaker, and \textbf{deps} a set of typed
links (\texttt{derived\_from} or \texttt{coupled\_with}) to other units.
Units persist across sessions in a \texttt{StateStore}. Where the data
carries session dates, the encoder stamps each unit it creates with the
turn's date; date handling is metadata-level throughout, so prompts are
unchanged on undated data. While extracting, the \texttt{TurnEncoder} may
also mark existing units as at risk of being \textbf{superseded} and
volunteer replacements.

\subsection{Update}
The update stage performs two operations. First, the \texttt{StateStore}
applies the flagged supersessions, setting the targeted unit's status to
\texttt{superseded}, and any volunteered replacement is added as a new active
unit. Second, a deterministic \texttt{Rechecker} traverses the dependency
graph $G = (U, E)$, where an edge $(u_i, u_j) \in E$ records that $u_i$
depends on $u_j$. For each $u_i$ with an edge to a unit whose status
\emph{just} changed, the \texttt{Rechecker} sets $u_i$ to
\texttt{needs\_recheck}, marking its content as possibly stale.

Note that a \texttt{superseded} unit stays in
the store for auditability but is inactive and withheld from the next stage,
while a \texttt{needs\_recheck} unit stays active but flagged. Because
dependencies are recorded at ingestion, the whole step runs in $O(|E|)$ and
adds no LLM calls.

\subsection{Test Time}
The \texttt{StateStore} deterministically assembles the valid state from all
active units, including those marked \texttt{needs\_recheck}, and renders it
as a structured block $\mathcal{S}$ grouped by priority and source, with each
flagged unit shown alongside the trigger that flagged it. Dated units render
with their dates, which double as recency markers, and where the benchmark
supplies a question date it is included as a ``today'' anchor.

Answering costs one further LLM call, over $\mathcal{S}$, the question, and a
prompt guiding recomputation of the current state
(Appendix~\ref{app:state-injection-prompts}). This deterministic layer decides which units are relevant and which are
stale. The LLM only recomputes the answer given those flags.

\definecolor{hcgray}{gray}{0.90}

\begin{table}
\centering
\small
\setlength{\tabcolsep}{5pt}
\renewcommand{\arraystretch}{1.12}
\caption{State-agreement accuracy (gold rate) on \textsc{StateMemBench}:
 $190$ in short scenarios ($\sim$165 turns, one probe
each) and $132$ in $44$ long fused scenarios ($\sim$600 turns, three
probes each), graded by a fixed
deepseek-v4-pro judge. Memory systems (Mem0 \citep{mem0}, A-Mem \citep{xu2025amemagenticmemoryllm},
LightMem \citep{fang2025lightmem}, MemoryOS \citep{kang-etal-2025-memory}, StateMem),
retrieval baselines (BM25, text-embedding-3-small \citep{openai_embed3small}), and
graph-RAG systems (nano-graphrag, HippoRAG, LightRAG, GraphRAG) run on each
substrate (Qwen-3.5-9B \citep{qwen2026qwen35}, thinking off; deepseek-v4-flash
\citep{xu2026deepseek}, thinking off); long-context baselines are raw
models with full history in context. StateMem ablations (indented)
excluded from best-marking. Per column within each panel, best \textbf{bold},
second best \underline{underlined}. \textsuperscript{***}StateMem's overall gold rate beats the strongest memory baseline on both backbones (paired McNemar test; both p < 0.001)}
\label{tab:state-mem-bench}
\begin{tabular}{c l c c c}
\toprule
& \textbf{Condition} & \textbf{Overall} & \textbf{Short} & \textbf{Long} \\
\midrule
\multirow{4}{*}{\rlab{Long-context}}
 & Qwen-3.5-9B        & \underline{0.149} & \underline{0.137} & 0.167 \\
 & Qwen-3.6-35B-A3B   & 0.097 & 0.063 & 0.146 \\
 & GPT-5.4-Nano       & \textbf{0.277} & \textbf{0.311} & \textbf{0.227} \\
 & DeepSeek-V4-Flash  & \underline{0.149} & 0.132 & \underline{0.174} \\
\midrule
\multirow{14}{*}{\rlab{Qwen-3.5-9B}}
 & No memory          & 0.006 & 0.000 & 0.015 \\
\cmidrule(l){2-5}
 & BM25               & 0.125 & 0.095 & 0.167 \\
 & Dense & 0.130 & 0.105 & 0.167 \\
\cmidrule(l){2-5}
 & nano-graphrag      & 0.143 & 0.126 & 0.167 \\
 & HippoRAG           & 0.130 & 0.105 & 0.167 \\
 & LightRAG           & 0.115 & 0.089 & 0.152 \\
 & GraphRAG           & \underline{0.224} & \underline{0.216} & \textbf{0.235} \\
\cmidrule(l){2-5}
 & Mem0               & 0.149 & 0.116 & 0.197 \\
 & A-Mem              & 0.127 & 0.100 & 0.167 \\
 & LightMem           & 0.019 & 0.021 & 0.015 \\
 & MemoryOS           & 0.025 & 0.026 & 0.023 \\
\hcrow
 & \textbf{StateMem (ours)} & \textbf{0.233}\textsuperscript{***} & \textbf{0.237} & \underline{0.227} \\
 & \abl{-- extraction-only}   & 0.155 & 0.126 & 0.197 \\
 & \abl{-- + supersession}    & 0.171 & 0.137 & 0.220 \\
 & \abl{-- w/o dep.-propagation} & 0.222 & 0.202 & 0.250 \\
 & \abl{-- w/o recompute-guid.}  & 0.208 & 0.195 & 0.227 \\
\midrule
\multirow{14}{*}{\rlab{deepseek-v4-flash}}
 & No memory          & 0.003 & 0.000 & 0.008 \\
\cmidrule(l){2-5}
 & BM25               & 0.143 & 0.105 & 0.197 \\
 & Dense & \underline{0.205} & 0.168 & \underline{0.258} \\
\cmidrule(l){2-5}
 & nano-graphrag      & 0.183 & \underline{0.174} & 0.197 \\
 & HippoRAG           & 0.184 & 0.132 & \underline{0.258} \\
 & LightRAG           & 0.093    & 0.058    & 0.144    \\
 & GraphRAG           & 0.174 & 0.137 & 0.227 \\
\cmidrule(l){2-5}
 & Mem0               & 0.177 & 0.168 & 0.189 \\
 & A-Mem              & 0.199 & 0.158 & \underline{0.258} \\
 & LightMem           & 0.012    & 0.021 & 0.000    \\
 & MemoryOS           & 0.025 & 0.011 & 0.045 \\
\hcrow
 & \textbf{StateMem (ours)} & \textbf{0.363}\textsuperscript{***} & \textbf{0.411} & \textbf{0.295} \\
 & \abl{-- extraction-only}   & 0.174 & 0.153 & 0.205 \\
 & \abl{-- + supersession}    & 0.298 & 0.337 & 0.242 \\
 & \abl{-- w/o dep.-propagation} & 0.373 & 0.400 & 0.333 \\
 & \abl{-- w/o recompute-guid.}  & 0.301 & 0.326 & 0.265 \\
\bottomrule
\end{tabular}
\end{table}

\section{Results}
We show \textsc{StateMem}'s competitive performance on existing memory benchmarks and evaluate \textsc{StateMemBench} on long-context LLMs, existing memory systems, retrieval baselines, and \textsc{StateMem}. We report the full results in Table \ref{tab:state-mem-bench}; for a brief overview of each memory system evaluated and further implementation details, see \S\ref{app:evaluation}. Each reported number is a single run per configuration: all answer and judge models decode at temperature~0 with reasoning traces disabled.

\subsection{Existing Memory Benchmark Results}
\label{sec:results-lme}
Among the systems we evaluate, StateMem scores highest of all \emph{memory systems} on LongMemEval on both substrates (0.580 on Qwen-3.5-9B and 0.656 on DeepSeek-V4-Flash; Mem0 is next at 0.566/0.594), within a point of the long-context baseline on DeepSeek (0.666); and on LoCoMo (0.566/0.592) it leads all memory systems and is level with long-context on DeepSeek (0.592 vs.\ 0.587) (Table~\ref{tab:lme-locomo}, with full version including question types in Table~\ref{tab:lme-locomo-qtype}).

The per-question-type numbers indicate where the gain comes from. Both benchmarks include question types that require tracking an update, and StateMem's margins are concentrated on those: temporal reasoning on LongMemEval (0.624 vs.\ 0.391 for long-context on deepseek, 0.398 vs.\ 0.248 on Qwen) and knowledge update (0.795 on deepseek). On single-hop recall and aggregation questions, where having the full history in context is hard to beat, StateMem comes within a few points of
long-context while reading only a bounded state. We therefore conclude state tracking does not cost recall.

\subsection{\textsc{StateMemBench} Results}
\label{sec:smb-results}
On StateMemBench, long-context ceases to be the strong baseline it is on recall benchmarks: even the best long-context model (GPT-5.4-Nano) reaches only 0.277 overall, and same-backbone long-context scores just 0.149 on both DeepSeek-V4-Flash and Qwen-3.5-9B. The adversarial traps defeat simply holding the full transcript in context.

Within each backbone panel, StateMem is the strongest condition. On DeepSeek it reaches 0.363, beating same-backbone long-context by 2.4$\times$ (vs.\ 0.149), the best memory system by 1.8$\times$ (A-Mem, 0.199), and the best baseline of \emph{any} type by a similar factor (Dense retrieval, 0.205). On Qwen, StateMem (0.233) leads the best memory system by 1.6$\times$ (Mem0, 0.149) and beats long-context (0.149), with GraphRAG (0.224) statistically level (McNemar over the 322 shared probes: 38 vs.\ 35 discordant, $p{=}0.82$). Explicit state tracking thus separates decisively from graph-structured retrieval on the stronger backbone, where the drift-rate itself falls ($-15$\,pp, \S\ref{sec:error}); on the weaker, the gain comes mainly from converting off-pool non-answers into in-pool answers. This is consistent with \S\ref{sec:error}'s finding that a capable answerer is needed to convert an explicit state representation into correct answers. We analyze \textit{why} the memory baselines underperform (as well as why StateMem largely does not) in \S\ref{sec:error}.

Category ordering holds across substrates. Anti-trap sits near ceiling almost everywhere (StateMem $\approx$0.87 on DeepSeek), since its gold answer is the anchored value. The compositional modes separate the systems---baselines are near zero on salience, sequence, and compound, and StateMem recovers most of the last two. Salience appears to be a substrate limit rather than a memory-design one, since every Qwen arm scores ${\leq}0.18$ against $0.569$ for GPT-5.4-Nano.

\paragraph{Which components pay their way?} Extraction alone recovers little (0.174 / 0.155); supersession marking is the largest single step (0.174$\to$0.298 on DeepSeek), and dropping recompute guidance costs 6\,pp. On the other hand, dependency propagation helps its target modes but over-propagates on Set~B anti-traps ($-12.5$\,pp), and removing it leaves DeepSeek slightly better (0.373 vs.\ 0.363). Again, since StateMem mirrors the policy family behind the traps (\S\ref{sec:taxonomy}), we read these margins as an upper bound and \S\ref{sec:results-lme} as a test of its generalization.

\subsection{Why Do Memory Systems Drift?}
\label{sec:error}
We utilize the closed-pool design discussed in \S\ref{sec:scenario_generation} to better understand the state tracking behavior of the evaluated models.
We calculate \textbf{drift-rate}, or how often an incorrect answer was the selected drift target (drift\,/\,$n$), where a lower rate is better, and report those results in Table~\ref{tab:drift_main}. Reading it requires separating two modes of failure: (1) when the method is actively anchoring on a stale value (an in-pool drift); and (2) when the method fails to surface a recognizable or plausible option at all (off-pool). Methods that answer predominantly off-pool (no memory, LightMem, and MemoryOS, which land off-pool on 60--94\% of cells) post deceptively low drift-rates by construction, not because they resist the drift target trap but because they rarely produce an in-pool (plausible) answer. We therefore report in-pool count (I-P) alongside drift-rate, and give full closed-pool breakdowns in Appendix Tables~\ref{tab:drift_full_ds} and~\ref{tab:drift_full_qwen}.

\begin{table}[h]
\centering
\small
\setlength{\tabcolsep}{6pt}
\renewcommand{\arraystretch}{1.12}
\caption{Accuracy on LongMemEval (full set, $n{=}500$, $k{=}20$) and LoCoMo
($n{=}1{,}985$), with each method run on both substrates (Qwen-3.5-9B;
deepseek-v4-flash; thinking off) and all answers graded by a fixed
deepseek-v4-pro judge. Per column, best in
\textbf{bold}, second best \underline{underlined}.}
\label{tab:lme-locomo}
\begin{tabular}{l cc cc}
\toprule
& \multicolumn{2}{c}{\textbf{LongMemEval}} & \multicolumn{2}{c}{\textbf{LoCoMo}} \\
\cmidrule(lr){2-3}\cmidrule(lr){4-5}
\textbf{Condition} & Qwen-9B & DS-V4 & Qwen-9B & DS-V4 \\
\midrule
Long context & 0.550 & \textbf{0.666} & \textbf{0.612} & \underline{0.587} \\
No memory    & 0.062 & 0.056 & 0.010 & 0.005 \\
\cmidrule(l){1-5}
Mem0         & \underline{0.566} & 0.594 & 0.481 & 0.462 \\
A-Mem        & 0.520 & 0.510 & 0.266 & 0.279 \\
LightMem     & 0.064 & 0.056 & 0.027 & 0.020 \\
MemoryOS     & 0.080 & 0.072 & 0.040 & 0.031 \\
\hcrow
\textbf{StateMem (ours)} & \textbf{0.580} & \underline{0.656} & \underline{0.566} & \textbf{0.592} \\
\bottomrule
\end{tabular}
\end{table}
Among methods that do engage the closed pool, the weak backbone (Qwen-3.5-9B) drifts heavily and near-uniformly: drift-rate sits at 61--66\% for long-context, Mem0, BM25, and StateMem alike; so on a weak answerer the memory layer barely changes the outcome. The methods separate on the stronger backbone (DeepSeek-V4-Flash). StateMem's drift-rate falls the most, from 64.3\% to 49.1\% ($-15$\,pp), and its correct-answer count rises the most, from 75 to 117 ($+42$): a capable answerer \textbf{converts would-be drifts into gold} once the state representation makes the operative value explicit. Long-context, Mem0, and BM25 barely move (drift-rate shifts by 1--3\,pp, correct answers by 0--9), indicating their bottleneck is \textbf{the memory layer, not the answerer}. StateMem is also the most engaged answerer on both backbones, with the fewest off-pool answers (32 on Qwen, 36 on DeepSeek) and the highest in-pool count, indicating the accuracy is not due to higher abstention.

\begin{table}
\centering
\small
\caption{Closed-pool ($n{=}322$) drift-rate and in-pool engagement, methods that produce non-trivial in-pool counts on at least one backbone. \textit{In-pool} (I-P) = method produced one of the closed-pool options. \textit{Drift-rate} (D-R) = drift\,/\,$n$ (errors are 0 for every method). StateMem's I-P is \textbf{bold}.}
\label{tab:drift_main}
\begin{tabular}{lrrrr}
\toprule
 & \multicolumn{2}{c}{Qwen-3.5-9B} & \multicolumn{2}{c}{DeepSeek-V4-Fl.} \\
\cmidrule(lr){2-3} \cmidrule(lr){4-5}
Method & I-P & D-R ($\downarrow$) & I-P & D-R ($\downarrow$) \\
\midrule
full context & 263 & 65.5 & 258 & 64.0 \\
Mem0 & 254 & 62.7 & 268 & 59.9 \\
BM25 & 243 & 60.9 & 244 & 59.9 \\
LightMem & 109 & 32.0 & 52 & 14.9 \\
\midrule
\rowcolor{hcgray}
\textbf{StateMem (ours)} & \textbf{290} & 64.3 & \textbf{286} & 49.1 \\
\bottomrule
\end{tabular}
\end{table}
\subsection{StateMem as a Wrapper}
\label{sec:wrapper}
To test the claim that state tracking is a separable axis, we build
\textsc{StateMemWrapper}: StateMem's approach to state as a prompt-level
transformation of the answer call, with none of its machinery. The
wrapper receives the transcript, the backend's retrieved chunks, and the
question, and replaces the answer call the backend would have made
anyway, so it adds no LLM calls.

The call fuses two stages. \textbf{Trace} writes, scoped by question, the
value chain of every slot the question touches---initial value, each
revision, and current operative value, in chronological order with turn
numbers---plus the latest stated inputs of every standing rule.
\textbf{Resolve} then commits under four precedence rules targeting the
failure modes of \S\ref{sec:prevalence}: later supersedes earlier,
standing rules outrank instances, derived quantities are recomputed
rather than quoted, and a fact is retired only by explicit supersession
or expiry. The trace is capped at 250 words and committed before the
answer, which blocks post-hoc justification. Overhead over the matched
control is a fixed 155-token instruction and that ${\leq}250$-word
trace, both inside the one call
(Appendix~\ref{sec:wrapper-cost}).

The control, \textsc{wrapper-ctrl}, matches the wrapper on call count,
word budget, answer format, and context; its first stage is generic
question-conditioned extraction, with no chains, no turn numbers, and no
resolution rules. $\Delta(\text{wrapper}-\text{wrapper-ctrl})$ therefore
isolates structure from sheer text volume. Both prompts appear verbatim
in Appendix~\ref{app:wrapper-prompts}.


\begin{table}
\centering
\small
\setlength{\tabcolsep}{4pt}
\caption{\textbf{StateMemWrapper composition sweep}; each backend ingests with its
unmodified pipeline and is evaluated on the same frozen paired $n{=}60$ set
per benchmark (StateMemBench: 30 short + 30 long scenarios, $k{=}10$;
LongMemEval: $k{=}20$) under three answer conditions: the backend \emph{alone}, a length- and
cost-matched generic control (\emph{+Ctrl}), and our state-tracing wrapper
(\emph{+SMW}). All answers are scored by a fixed deepseek-v4-pro judge.
Best condition per row in \textbf{bold}. Dense on LongMemEval/Qwen has
$n{=}59$ (one case failed at ingest).}
\label{tab:wrapper}
\begin{tabular}{cl cc>{\columncolor{hcgray}}c cc>{\columncolor{hcgray}}c}
\toprule
& & \multicolumn{3}{c}{\textbf{Qwen-3.5-9B}} & \multicolumn{3}{c}{\textbf{deepseek-v4-flash}} \\
\cmidrule(lr){3-5}\cmidrule(lr){6-8}
& \textbf{Backend} & Alone & +Ctrl & +SMW & Alone & +Ctrl & +SMW \\
\midrule
\multirow{6}{*}{\rlab{StateMemBench}}
 & Mem0      & 25.0 & 28.3 & \textbf{56.7} & 28.3 & 56.7 & \textbf{71.7} \\
 & A-Mem     & 23.3 & 35.0 & \textbf{56.7} & 33.3 & 51.7 & \textbf{75.0} \\
 & LightMem  &  3.3 & 30.0 & \textbf{61.7} &  1.7 & 51.7 & \textbf{68.3} \\
 & MemOS     &  5.0 & 25.0 & \textbf{56.7} &  1.7 & 50.0 & \textbf{68.3} \\
 & BM25      & 20.0 & 31.7 & \textbf{56.7} & 21.7 & 48.3 & \textbf{70.0} \\
 & Dense     & 21.7 & 28.3 & \textbf{56.7} & 31.7 & 43.3 & \textbf{70.0} \\
\midrule
\multirow{6}{*}{\rlab{LongMemEval}}
 & Mem0      & 38.3 & 36.7 & \textbf{48.3} & 45.0 & 58.3 & \textbf{60.0} \\
 & A-Mem     & 41.7 & 31.7 & \textbf{56.7} & 35.0 & 51.7 & \textbf{55.0} \\
 & LightMem  &  5.0 & 11.7 & \textbf{28.3} &  3.3 & 31.7 & \textbf{35.0} \\
 & MemOS     &  6.7 &  8.3 & \textbf{23.3} &  6.7 & 31.7 & \textbf{36.7} \\
 & BM25      & 16.7 & 16.7 & \textbf{36.7} & 16.7 & 40.0 & \textbf{41.7} \\
 & Dense     & 47.5 & 32.2 & \textbf{50.8} & 53.3 & \textbf{61.7} & 56.7 \\
\bottomrule
\end{tabular}
\end{table}

\subsubsection{Evaluation}
We evaluate all six backends on both substrates against frozen paired
$n{=}60$ sets from \textsc{StateMemBench} and LongMemEval
\citep{wu2024longmemeval} (Table~\ref{tab:wrapper}). Each store is
built once and answered three ways, so \emph{alone}, \emph{+Ctrl}, and
\emph{+SMW} face identical retrieved chunks under a single fixed judge.


The wrapper improves every backend on both benchmarks. On
\textsc{StateMemBench} it adds $+31.7$ to $+66.6$ points over each backend alone (largest for the near-floor LightMem and MemOS), of which structure contributes $+15.0$ to $+31.7$. This is significant in all twelve cells (paired McNemar $p{\leq}0.04$, bootstrap CIs exclude zero) and exceeds the control in 23 of 24. On LongMemEval the split is
substrate-dependent, as structure carries the gain on Qwen ($+11.7$ to $+25.0$; the control alone often hurts), while on the DeepSeek substrate the control recovers most of the lift and structure adds only $-5$ to $+5$. This implies a strong answerer has less need for the state structure. Additionally, a question-blind 250-word summary falls below the no-wrapper baseline on both benchmarks and carries the highest drift rate of any arm, showing the importance of question conditioning.

Notably, the wrapper outperforms \textsc{StateMem} itself on the same scenarios (0.283 vs.\ ${\geq}0.567$ on Qwen). When the transcript fits in context, resolving state at answer time is good enough, but the persistent store can be useful when it does not fit. Prompts, backend protocol, blind-summary ablation, and cost analysis are
in Appendix~\ref{app:wrapper}; per-cell tables are
Tables~\ref{tab:wrapper-split-ourbench}
and~\ref{tab:wrapper-lme-transitions}.
\section{Conclusion}
We analyze state tracking as a capability distinct from recall and reasoning, and show that state drift is already present, though unmeasured, in the failure states of existing benchmarks even under perfect retrieval. We introduce
\textsc{StateMemBench}, which evaluates state tracking through 234
multi-session scenarios spanning five failure modes (status, salience,
sequence, compound, and anti-trap) across three domains, and find that existing
memory systems and long-context baselines struggle across all of them. We then
present \textsc{StateMem}, a state-first method representing state and its
relational dependencies, which improves over the strongest same-backbone
baseline by up to 1.8$\times$ on \textsc{StateMemBench}, surpasses
same-backbone long-context there, and stays competitive with it on recall
benchmarks; ablations attribute the gain chiefly to supersession marking and
recompute prompting. Finally, the state-tracking axis is separable: applied as an answer-time
wrapper, it lifts every backend we test, with a matched
control isolating the structural share. Current memory systems lack just this explicit state discipline.

\section*{Acknowledgments}
Research was supported in part by the AI Institute for Molecular Discovery, Synthetic Strategy, and Manufacturing: Molecule Maker Lab Institute (MMLI), funded by U.S. National Science Foundation under Award 2505932, NSF IIS 25-37827, and the Institute for Geospatial Understanding through an Integrative Discovery Environment (I-GUIDE) by NSF under Award No. 2118329.  The research has used the Delta/DeltaAI advanced computing and data resource, supported  in part by the University of Illinois Urbana-Champaign and through allocation \#250851 from the Advanced Cyberinfrastructure Coordination Ecosystem: Services \& Support (ACCESS) program, which is supported by National Science Foundation grants OAC 2320345, \#2138259, \#2138286, \#2138307, \#2137603, and \#2138296.  Any opinions, findings, and conclusions or recommendations expressed herein are those of the authors and do not necessarily represent the views, either expressed or implied, of DARPA or the U.S. Government. Miri Liu was supported by the Amazon AI PhD Fellowship.

\bibliography{aaai2027}

\newpage
\appendix

\clearpage

\section{State Drift Labeling and Validation Details}
\label{app:drift-label-validation}

\subsection{Buckets and labeling procedure}
\label{app:buckets}
Each confirmed failure is one $(case, turn, slot)$ point assigned to
exactly one bucket. An ordered rule stops at the first match: (Q1) is the
gold fact in context? (no $\to$ retrieval failure) (Q2) is the gold well
defined? (no $\to$ gold issue; $\leq3\%$ everywhere, omitted from
Table~\ref{tab:cross-bench-failure}) (Q3) does the output engage the slot
domain? (no $\to$ comprehension failure) (Q4) is it correct but wrongly
formatted? ($\to$ schema mismatch) (Q4.5) where the gold may be an
abstention, did the agent confabulate an in-domain value? ($\to$ failed
abstention; vacuous where gold is always concrete) (Q5) does the output
surface the current value but err in the inference? ($\to$ reasoning
failure). Only a point surviving all six reads, whose output matches an
available but stale or wrong-scope version of a never-retracted fact, is
labeled drift; points whose evidence supports no assignment are dropped,
not defaulted. The never-retracted condition excludes legitimate update:
the judge must cite evidence that the newer version was still in force,
so an agent following a genuine revision is never charged. Q1 is resolved
mechanically where possible: in shopping and search the gold token is in
the printed candidate set by construction; in travel a prompt scan
resolves $99.1\%$ of cases; on LongMemEval oracle availability is
guaranteed by the prompt. Q2--Q5 are judge-applied.

\paragraph{Notes on Table~\ref{tab:cross-bench-failure}.} Rows reach
$100\%$ together with an omitted gold-issue share ($\leq3\%$). MA rows
pool $8$ memory variants after the adversarial filter. ``FA'' is
``--'' where the benchmark's gold is never an abstention. The LME row
reports the cross-family confirmed drift rate with the remaining cells
from the conservative adjudication (\S\ref{app:judging}); its drift
share carries a Wilson $95\%$ CI of $[30\%, 60\%]$ ($n{=}36$). The
$\tau^2$-Z row is descriptive: drift CI $[39\%, 82\%]$, and its
$0.0\%$ cells are indistinguishable from non-observation.

\subsection{Conservatism: adversarial filter and cross-family confirmation}
\label{app:judging}
The adversarial filter, a second pass that attempts to re-classify every
drift candidate as another bucket, downgrades $32$--$47\%$ of candidates
across benchmarks, so surviving labels are conservative by construction.
Judging proceeds in two stages. First, two independent decodes of the
same DeepSeek-V4 judge (temperature $0$, identical decision-tree prompt)
label every confirmed failure; their agreement is
Table~\ref{tab:agreement}. The twelve binary disagreements are structured: GPT-4o pulls seven DeepSeek-drift cases (six of them temporal-reasoning, date-format-adjacent) out to schema mismatch or failed abstention, and pushes four abstention-gold cases into drift; the conservative rule takes the non-drift side of every flip. Second, on LongMemEval oracle (the benchmark
carrying the perfect-retrieval claim) we add a judge from a different
model family (GPT-4o, same rubric and prompt) and report the
\emph{cross-family confirmed} rate: a failure counts as drift only if
both families say drift. The DeepSeek pass yields $63.9\%$ drift and
GPT-4o $55.6\%$; the confirmed rate is $44.4\%$ ($16/36$), which is what
the paper reports. Where exactly one family says drift we take the
non-drift reading, giving the conservative-adjudicated row in
Table~\ref{tab:cross-bench-failure}. Cross-family binary agreement is
$69.4\%$ raw ($\kappa{=}0.37$, PABAK${=}0.39$), lower than same-family
agreement. This is why we report the confirmed rate rather than either
family's rate alone.

\subsection{Inter-judge agreement}
\label{app:inter-judge}
Cohen's $\kappa$ deflates mechanically when one class dominates the
marginals; PABAK ($=2P_a-1$, recomputable from raw agreement alone) and
Gwet's AC1 correct for this.

\begin{table}[t]
\centering\footnotesize
\setlength{\tabcolsep}{3.5pt}
\begin{tabular}{lrrrrrr}
\toprule
Benchmark & $n$ & agr (\%) & $\kappa$ & PABAK & AC1 & $\kappa_{\text{mode}}$ \\
\midrule
MA-shopping & 179 & 91.6 & 0.17 & 0.83 & 0.91 & 0.00 \\
MA-search   & 110 & 72.7 & 0.31 & 0.45 & 0.55 & 0.09 \\
LME oracle  &  36 & 77.8 & 0.67 & 0.56 & 0.56 & 0.01 \\
\midrule
LME oracle (cross) & 36 & 69.4 & 0.37 & 0.39 & --- & --- \\
\bottomrule
\end{tabular}
\caption{Inter-judge agreement on the binary drift decision: two
same-model DeepSeek-V4 decodes (top), and DeepSeek-V4 $\times$ GPT-4o
(bottom row cross family). MA-travel is omitted ($2$ doubly-labeled cases).
MA-shopping is the one cell where $\kappa$ misrepresents reliability:
both passes call ${>}90\%$ of failures drift \emph{pre-filter}, which
crushes the chance correction despite $91.6\%$ raw agreement.}
\label{tab:agreement}
\end{table}

Mode $\kappa$ (which drift sub-mechanism) is near zero on every benchmark
($0.00$--$0.09$) while binary $\kappa$ is fair-to-substantial: judges
agree on \emph{whether} a failure is drift but not on \emph{why the wrong
fact won}. This is the empirical basis for \S\ref{sec:prevalence}'s
argument that mechanism must be fixed by construction rather than
inferred post hoc.

\subsection{Human anchoring}
\label{app:human-anchor}
Judge--judge agreement cannot establish that the categories are real, so
we anchored the rubric to two human annotators under two protocols.
\emph{Track A (structured, $n{=}12$)}: annotators answer fixed
sub-questions (was the gold fact in context; did the agent reference it
within $K{=}3$ turns; was the referenced scope correct), then record a
free observation from a neutral menu; the bucket is \emph{derived} from
these answers rather than chosen, so the taxonomy cannot be imposed by
the form. \emph{Track B (blind, $n{=}6$)}: annotators describe the
failure in free text \emph{before} seeing the taxonomy; a researcher then
maps descriptions to buckets post hoc. The sample is stratified across
LME-oracle, MA-shopping, and MA-search and oversamples judge-disputed
cases ($6/12$ in Track A), making it a deliberately hard audit set.

\emph{Inter-annotator reliability.} On Track A the annotators agree on
the binary drift decision at $83.3\%$ raw ($\kappa{=}0.43$,
PABAK${=}0.67$); on the full bucket label, $50.0\%$ raw
($\kappa{=}0.42$). On Track B the neutral surface-observation label
reaches $\kappa{=}0.80$ ($83.3\%$ raw): blind to the taxonomy, annotators
see the same failure surface. All $12$ blind free-text descriptions
($6$ cases $\times$ $2$ annotators) map onto existing buckets; no
description required a novel category, and one annotator's blind
description of an LME case independently reconstructs the salience
mechanism (``the agent is just grepping for mentions'').

\emph{Human vs.\ judge.} Binary agreement with the judge is $58.3\%$ and
$41.7\%$ for the two annotators, and both assign \emph{less} drift than
the judge ($8.3\%$ and $25.0\%$ vs.\ $50.0\%$ on this disputed-enriched
sample). The departures are replicated rather than idiosyncratic: on five
of twelve cases both annotators converge on the same alternative label,
concentrated in two patterns. (i) \emph{Wrong slot vs.\ drift}: answers
returning a person where gold is a date/year, which the judge's
\emph{other\_drift} mode absorbs and both humans call schema mismatch;
the cross-family confirmation rule of \S\ref{app:judging} corrects the
same boundary. (ii) \emph{Equivalence}: answers the judge calls schema
mismatch that gold's own notes accept as alternate forms (a listed
alternate name; an omitted legal suffix), which both humans call
not-a-failure. Both patterns are policy boundaries, not disputes about
the drift category: the human audit and the cross-family confirmation
converge on the same correction, applied before any number is reported.
The annotation kit (forms, keys, and the analyzer) is released with the
benchmark.

\subsection{LongMemEval failure anatomy}
\label{app:lme-anatomy}
Of $306$ oracle questions, DeepSeek-V4-Flash fails $36$. The
conservative-adjudicated distribution is $44.4\%$ state drift ($16$),
$27.8\%$ failed abstention ($10$), $25.0\%$ schema mismatch ($9$),
$2.8\%$ reasoning ($1$), $0\%$ comprehension and retrieval (recall
$=1.0$ by construction). By question type, failures of multi-session
questions are drift at $71.4\%$ ($10/14$) vs.\ $25.0\%$ ($2/8$) for
knowledge-update, identical under the single-family and cross-family
confirmed labelings.


\subsection{Reasoning-vs-state separation A/B}
\label{app:reasoning-ab}
If drift were reasoning shortfall in disguise, a stronger reasoner would
close the gap. We test this directly on LongMemEval multi-session oracle
($n{=}50$ paired items, identical prompts, full history in context so no
retrieval confound): the only manipulation is enabling
DeepSeek-V4-Flash's native reasoning trace, with the judge held constant
across arms.

\begin{table}[h]
\centering\small
\begin{tabular}{lrr}
\toprule
Arm & correct & rate \\
\midrule
thinking off & 42/50 & 84.0\% \\
thinking on  & 38/50 & 76.0\% \\
\midrule
$\Delta$ & & $-8.0$\,pp \\
\bottomrule
\end{tabular}
\quad
\begin{tabular}{lr}
\toprule
Transition & $n$ \\
\midrule
off-fail $\to$ on-pass & 3 \\
off-pass $\to$ on-fail & 7 \\
both pass & 35 \\
both fail & 5 \\
\bottomrule
\end{tabular}
\caption{Paired reasoning A/B. Reasoning does not close the gap: the
$-8.0$\,pp point estimate rests on $3$ rescues against $7$ new failures,
which does not establish a direction (exact McNemar $p{=}0.34$). The
off-pass\,$\to$\,on-fail cell includes the case dissected in
Appendix~\ref{app:drift-case}.}
\label{tab:reasoning-ab}
\end{table}

The dispositive contrast is the asymmetry, not the point estimate:
reasoning leaves the failure set where it was, while state-maintenance
interventions move it (the wrapper lifts the same six backends by
$+20.9$\,pp mean over each backend alone on this task and backbone,
with the structure share decomposed by the matched control in
\S\ref{sec:wrapper}). These failures lie in state construction, not
reasoning capacity.


\section{Drift Case}
\label{app:drift-case}

One LongMemEval-oracle case (\texttt{46a3abf7}, multi-session), chosen
because it is cross-family confirmed and both judge families
independently assign not only \emph{drift} but the same mechanism
(salience). All sessions are in the prompt; recall is $1.0$.

\paragraph{What the context establishes.} Across three sessions the user
has three tanks on record:

\begin{itemize}\setlength\itemsep{1pt}
\item \textbf{May 21}, mid-session, inside a question about plant
temperatures: \emph{``I've also been taking care of a small 1-gallon
tank that I set up for a friend's kid, which has a few guppies\ldots''}
Mentioned once, never again.
\item \textbf{May 23}: \emph{``I have a 5-gallon tank with a solitary
betta fish named Finley''}; the same session announces \emph{``I've
since set up a new 20-gallon community tank.''}
\item \textbf{May 27}: a full session about the 20-gallon,
\emph{``I've finally set up my 20-gallon freshwater community tank,
which I've named `Amazonia'\,''}, followed by a dozen turns on its
fish, food, algae, plants, and driftwood.
\end{itemize}

\paragraph{The probe (May 30).} \emph{``How many tanks do I currently
have, including the one I set up for my friend's kid?''} Gold:
\textbf{3}. Agent: \textbf{2}.

\paragraph{Why this is drift and nothing else.} The question itself
names the quiet fact, and the agent still fails to count it. Every
alternative reading is closed: not retrieval (oracle: every session is
in the prompt), not comprehension (the agent returns a well-formed
count), not schema (a number is a number), and not a legitimate update
(no tank was ever given away or retracted). Nothing in the output
engages the 1-gallon tank. Nor is it reasoning starvation: in the
paired A/B of Appendix~\ref{app:reasoning-ab} this same case flips---
without the reasoning trace the model answers $3$ (correct); with
deliberation enabled it answers $2$. What remains is the signature
asymmetry: the 20-gallon tank has a \emph{name} and two sessions of
repeated discussion; the 1-gallon tank has one clause inside a question
about plant care. The louder fact is fully attended; the quieter one,
six days and dozens of turns upstream, silently drops out of the state,
even when the probe points straight at it. Both judge families label
this \emph{state drift, salience mode}, independently and verbatim.
This is the mechanism \textsc{StateMemBench} manufactures by
construction: a quiet operative fact, a louder competitor, and a
decision that requires the quiet one.

\section{Benchmark Comparison}
\label{app:benchmark-comparison}
Table~\ref{tab:benchmark-comparison} scores each benchmark against the
requirements of state-tracking evaluation. The columns:
\begin{itemize}\setlength\itemsep{2pt}
\item \emph{Multi-session dialogue}: probes span temporally separated
conversational sessions, so state must persist across session boundaries
rather than within one context episode.
\item \emph{Scale}: reported size in each benchmark's native unit
(turns, sessions, or tasks).
\item \emph{Updates central}: state updates are the measured phenomenon,
not one question sub-type among many.
\item \emph{Supersession}: correct answers require later-overrides-earlier
resolution among multiple recorded values of the same fact.
\item \emph{Verifiable gold}: the gold answer derives from an executable
or programmatic state rather than from annotation (mapping a system's
free-form answer onto that key may still use a judge).
\item \emph{Drift scored}: the superseded value is a distinct scored
outcome, separating acting on old state from generic error.
\item \emph{Anti-update controls}: probes that reward retaining an
earlier value against a later, non-authoritative mention, so an
always-prefer-the-latest heuristic cannot score well.
\item \emph{Paired horizons}: the same probes appear at a short and a
long horizon where the \emph{operative update history} grows with
length, isolating maintenance from distractor volume.
\end{itemize}

\paragraph{LoCoMo \shortcite{locomo}.} Multi-session dialogues averaging
roughly $300$ turns over as many as $35$ sessions, with QA over facts and
events. Questions target recall, multi-hop composition, and temporal
ordering of \emph{static} facts. No question requires resolving a
superseded value, gold is free-text judged, and the stale value carries
no separate score. Its adversarial and unanswerable questions reward
declining when evidence is absent, which is not a control against
spurious updates.

\paragraph{LongMemEval \shortcite{wu2024longmemeval}.} Multi-session
dialogue with \emph{knowledge update} as one of six question types
(partial updates; those questions require preferring the later value,
which makes supersession partial). Gold is curated free text with no
programmatic verification, and the stale value is not a scored outcome.
Its abstention questions reward declining on missing evidence rather than
retaining a value against a spurious update, so they are not an
anti-update control. The $S$ and $M$ variants embed the same $500$
questions among more filler sessions ($\sim$40--80 vs.\ $\sim$500),
varying distractor volume while the update history itself is unchanged.
These are not paired state horizons.

\paragraph{MemoryAgentBench \shortcite{hu2025evaluating}.} Mixes
conversational and document-stream inputs, making multi-session dialogue
partial. Its FactConsolidation split orders counterfactual edit pairs so
that the new value follows the old, which makes updates a measured
sub-task with later-overrides-earlier resolution and makes part of the
gold programmatically constructed, partial on all three counts.
Evaluation is answer accuracy under an LLM judge, and the framework does
not separate an outdated answer from any other wrong answer, so the stale
value is not a scored outcome. There are no anti-update probes and no
paired horizons.

\paragraph{MemoryArena \shortcite{he2026memoryarena}.} Multi-session,
environment-grounded tasks whose interdependent subtasks are scored by
execution against the environment, so gold is verifiable. Its sessions
are agentic rather than conversational, which makes the multi-session
dialogue requirement partial. The measured phenomenon is task completion
under memory load: updates arise incidentally, no supersession chains are
constructed, and a failure is recorded as task failure rather than as
selection of a stale value.

\paragraph{STATE-Bench \shortcite{statebench2026}.} Enterprise task
episodes across customer support, travel, and shopping, each a
self-contained scenario over a pre-populated database with deterministic
assertions defining success. The state at issue is environment state
mutated by tool calls, not a tracked dialogue state, and each episode
stands alone rather than extending a multi-session history. Completion is
database-verified, but the reported evaluation also includes an
LLM-judged user-experience score, so verifiability is partial.
Its policy checks do supply a restricted anti-update control: the
simulated user can press for an action the database does not license, and
passing requires the agent not to act on that assertion. This gates
whether an action is permitted rather than which value of a fact remains
operative, so the control is partial. Supersession chains are not
constructed, and the stale value is not a distinct outcome.

\paragraph{StateMemBench (ours).} Here updates are the measured
phenomenon rather than a sub-type. Every probe is generated from an
executable state program, which makes the gold answer itself computed
rather than annotated; a judge enters only to map free-form answers onto
the closed pool. A deterministic word-boundary matcher decides the
$28\%$ of answers that name exactly one pool option verbatim and agrees
with the judge on $94.1\%$ of those; the judge maps the rest, which
name zero or several options.
The superseded value is a distinct closed-pool outcome, so drift is
scored rather than folded into generic error. Anti-update probes present
a later, non-authoritative mention of a value the agent should retain,
penalising an always-prefer-the-latest heuristic. And Set~A and Set~B
present identical probe structure at $\sim$165 and $\sim$600 turns, with
the operative update history itself lengthening rather than the
surrounding distractor volume, so length is the only thing that varies.

\section{StateMemBench}

\subsection{Research DOIs}
\label{app:dois}
The following DOIs are of the papers we used to ground our synthetic scenario generation for the research domain. No substantial verbatim text from these papers is reproduced, and we do not make any claims about the papers' factual accuracy. We intend for our generated scenarios and benchmark to be used purely for research purposes.

\begin{table}[h]
\centering
\small
\begin{tabular}{@{}ll@{}}
\toprule
\textbf{Source} & \textbf{DOI} \\
\midrule
Biomedical (bioRxiv)
  & \texttt{10.1101/482679} \\
  & \texttt{10.1101/512657} \\
\midrule
Computer Science (arXiv)
  & \texttt{10.48550/arXiv.2407.13248} \\
  & \texttt{10.48550/arXiv.2604.10985} \\
  & \texttt{10.48550/arXiv.2604.11543} \\
  & \texttt{10.48550/arXiv.2604.21253} \\
  & \texttt{10.48550/arXiv.2605.02392} \\
  & \texttt{10.48550/arXiv.2605.06196} \\
\bottomrule
\end{tabular}
\caption{Source DOIs used to ground synthetic scenario generation, grouped by domain.}
\label{tab:dois}
\end{table}
\subsection{StateMemBench: Research Example}
This is an \textbf{oracle} example scenario from StateMemBench (i.e., we include only turns that are relevant to the probe) from the research domain in the \textbf{salience} category. Speakers are named collaborators; the standing rule selects the model with the best F1 as the main-table headline.

\begin{tcolorbox}[colback=gray!5,colframe=black!60,title={Research / Salience}]
\small
\textbf{s1.t1 PI:} ``One standing call before we dive in: whatever ends up best on F1 goes in the main table, full stop.''\\[4pt]

\textbf{s3.t5 Engineer:} ``I was looking at the classifier numbers this morning: BERT Classifier is at 0.448, Logistic Regression baseline is at 0.338.''\\[4pt]

\textbf{s9.t6 Researcher:} ``Just so it's on the record --- BERT Classifier is still what I'm using as the reference point when I'm checking whether the stratified numbers look reasonable. Just the default workhorse for sanity checks.''\\[2pt]
\textbf{s12.t4 Researcher:} ``BERT Classifier is what I'd keep as the reference point while we sort out whether this is a data artifact or a genuine modeling gap.''\\[4pt]

\textbf{s15.t1 Researcher:} ``I was reviewing the latest numbers and, just so it's on the record, Logistic Regression baseline is at 0.695 on F1 now.''\\[4pt]

\textbf{Probe (after s15) --- PI:} ``Finalizing the main table this week --- which one's the headline row?''\\[2pt]
\textbf{Ground truth:} \emph{Logistic Regression baseline} (0.695; it was quietly revised past BERT Classifier, which stayed at 0.448 despite being repeatedly named as the team's ``workhorse.''). The \emph{drift} answer is BERT Classifier --- the salient, most-mentioned model.
\end{tcolorbox}

\subsection{StateMemBench: Shopping Example}

This is an \textbf{oracle} example scenario from StateMemBench (i.e., we include only turns that are relevant to the probe) from the shopping domain in the \textbf{sequence} category. The two-week total is a quantity \emph{derived} from the weekly amount, and must be recomputed after the weekly amount changes rather than re-quoted.

\begin{tcolorbox}[colback=gray!5,colframe=black!60,title={Shopping / Sequence}]
\small
\textbf{2.3 user:} ``Also going forward I want 10 packs of the snack packs each week, just set that as my regular amount.''\\[2pt]
\textbf{2.4 assistant:} ``Noted --- 10 packs weekly.''\\[4pt]

\textbf{11.7 user:} ``Snack packs --- ok so 10 a week, two weeks, that's 20 packs. So 20.''\\[2pt]
\textbf{11.8 assistant:} ``20 snack packs, done.''\\[4pt]

\textbf{15.7 user:} ``Snack packs --- going down to 7 packs a week btw --- so 7 today.''\\[2pt]
\textbf{15.8 assistant:} ``Noted. 7 snack packs added.''\\[4pt]

\textbf{Probe (after s15) --- user:} ``Placing the 2-week school-trip order --- how many packs total?''\\[2pt]
\textbf{Ground truth:} \emph{14} ($7 \times 2$; the weekly amount dropped to 7, so the two-week order recomputes). The \emph{drift} answer is 20 --- the earlier two-week total, stated aloud back when the weekly amount was 10 and never recomputed.
\end{tcolorbox}

\subsection{StateMemBench: Personal Finance Example}

This is an \textbf{oracle} example scenario from StateMemBench (i.e., we include only turns that are relevant to the probe) from the personal finance domain in the \textbf{status} category. A tiered standing rule maps the bonus amount to a destination; the probe follows a quarter in which the user emphatically committed to one tier before the amount changed.

\begin{tcolorbox}[colback=gray!5,colframe=black!60,title={Finance / Status}]
\small
\textbf{3.1 user:} ``Setting a rule for when my quarterly bonus lands: if it comes in under 1900, just leave it in checking; if it's between 1900 and 4000, top up the travel sinking fund; between 4000 and 6200 put it toward the car loan; and if it's over 6200, max the Roth contribution.''\\[4pt]

\textbf{7.1 user:} ``Bonus hit. Quarterly came in at \$4,883.''\\[2pt]
\textbf{7.2 assistant:} ``Got it. Put it toward the car loan, then.''\\[4pt]

\textbf{10.1 user:} ``The new quarterly bonus just posted --- \$7,017. I'm maxing the Roth, that's the plan, doing it this week.''\\[2pt]
\textbf{10.4 assistant:} ``Confirmed. I'll treat the Roth as fully funded once that \$3,500 clears.''\\[4pt]

\textbf{15.3 user:} ``Quarterly bonus came in, it's \$1,517 this time.''\\[2pt]
\textbf{15.4 assistant:} ``Got it.''\\[4pt]

\textbf{Probe (after s18) --- user:} ``With this quarter's bonus, I'll max the Roth contribution like last time, right?''\\[2pt]
\textbf{Ground truth:} \emph{leave it in checking} (the \$1,517 bonus falls in the lowest tier, so the rule re-resolves). The \emph{drift} answer is \emph{max the Roth contribution} --- the destination the user loudly committed to for the earlier \$7,017 bonus, now stale.
\end{tcolorbox}

\subsection{Scenario Generation Pipeline}
\label{app:scenario-gen}
\subsubsection{Domain Grounding}
\label{app:domains}
Every sampled program is \emph{grounded} with surface material from real data (data sources in \S\ref{sec:state-mem-bench_datasets}). 

\paragraph{Research}
We collect open-access research papers and reduce each to a cached \emph{fact sheet} of team roles, methods, datasets, and metrics from which scenario entities are drawn. See Appendix~\ref{app:dois} for paper details.

\paragraph{Shopping}
We derive consumer personas from the Instacart Online Grocery Basket Analysis Dataset \cite{instacart2017dataset}, sampling users for diversity over basket size, order cadence, reorder habit, and top department.

\paragraph{Personal Finance}
We derive personas from the Credit Card Transactions Dataset \citep{lee2024creditcard}, a synthetic credit-card transaction dataset, sampling card holders for diversity over age, income, debt, card count, and credit score.

\subsubsection{Rendering}
A strong LLM (\texttt{sonnet-4.6}) renders the symbolic event program into natural multi-session dialogue. The LLM is not given specific turns, but rather general instructions (e.g., ``in session 14 the user must state they have received a bonus of \$7540''). We also state facts or phrases the LLM must avoid in rendering. Since Set B scenarios are very long, we render in chunks with a running summary provided to the LLM at each new chunk. For shopping and finance, we render scenarios as dialogue between a user and helpful assistant, whereas for research, we render scenarios with multiple named speakers to better simulate memory in collaborative settings. Once scenarios are rendered, we programmatically verify that the load-bearing facts appear, and appear in the correct assigned session, as well as that no banned phrases appear in the final scenario. Any scenarios that fail are re-rendered until they pass.

\subsection{Validation}
\label{app:smb-validation}
All scenarios in the benchmark pass two filters. First, a strong LLM reader (\texttt{sonnet-4.6}), given only relevant sessions, must be able to recover the gold answer on $\geq 2/3$ of samples. Second, a weak LLM reader (\texttt{haiku-4.5}) with the full transcript should produce the drift answer instead on $\geq 3/5$ of samples (or the gold for anti-trap).

Two notes on this gate: the answerability threshold is $\geq 2/3$, not
$1.0$, so absolute scores sit against a ceiling somewhat below $1.0$;
and difficulty is defined against LLM readers by design, since the
benchmark measures whether a memory system protects its reader from
such traps. Gold and drift are computed before rendering, so the
renderer cannot alter trap semantics.

\subsection{Reasoning-vs-State Separation}
\label{app:reasoning-separation}

The reasoning-trace A/B referenced here (holding prompt, context, and
model fixed while toggling only the reasoning trace on $n{=}50$ paired
LongMemEval multi-session oracle cases) is reported in full, with its
per-case transition table, in
Appendix~\ref{app:drift-label-validation}
(Table~\ref{tab:reasoning-ab}).
\section{StateMem}
\label{app:prompts}
\subsection{TurnEncoder}
\label{app:turn-encoder-prompt}

\begin{tcolorbox}[smbprompt, title=Task Context]
\ttfamily\small
\setlength{\parskip}{4pt}
\setlength{\parindent}{0pt}
\#\# Task context:

You are tracking state for a multi-session user-assistant conversation. The user states constraints and preferences; later turns may supersede or update them. Track:

\hspace*{1em}-- hard constraints \\
\hspace*{1em}-- soft preferences \\
\hspace*{1em}-- entity status (active / closed / superseded) \\
\hspace*{1em}-- cross-unit dependencies
\end{tcolorbox}

\begin{tcolorbox}[smbprompt, title=TurnEncoder System Prompt]
\ttfamily\small
\setlength{\parskip}{4pt}
\setlength{\parindent}{0pt}
You are a state maintenance system for a task-executing agent. After each conversation turn, you update the agent's tracked state.

\#\# Current conversation state (from prior turns): \\
\{current\_state\}

\#\# Latest turn (turn \{turn\_number\}): \\
\{latest\_turn\}

\{task\_context\}

\#\# Your job: \\
Analyze the latest turn and determine what state updates are needed. You can:

1. ADD new units: New constraints, commitments, or facts that emerged this turn. \\
\hspace*{1.50em}- From USER statements: explicit requests, preferences, corrections \\
\hspace*{1.50em}- From TOOL results: new facts revealed by tool calls (e.g., membership level, flight status, reservation details, payment info) \\
\{action\_watching\_instructions\}

2. SUPERSEDE existing units: When new information contradicts or updates an existing unit. \\
\hspace*{1.50em}- Example: user changes payment preference from credit card to gift card \\
\hspace*{1.50em}- Example: tool result shows user is silver, not gold as claimed earlier \\
\hspace*{1.50em}- Include the ID of the unit being superseded. \\
\hspace*{1.50em}- Look ACTIVELY for supersession patterns: "actually", "instead", "no longer", "switched to", "moved off", "cancelled", "replaced X with Y", "now Y not X".

3. TRACK PROGRESS: If the user requested multiple actions (e.g., "downgrade all 5 reservations", "cancel these 3 flights"), track what has been completed and what remains. \\
\hspace*{1.50em}- Add a progress unit like: "Completed 2 of 5 reservation downgrades. Remaining: X7BYG1, EQ1G6C, BOH180" \\
\hspace*{1.50em}- Update this unit each turn as more items are completed. \\
\hspace*{1.50em}- Mark as "hard" priority so the agent doesn't forget remaining items.

4. LINK CASCADE / DERIVED dependencies: When a new unit's validity depends on another unit's status or value, surface that link. \\
\hspace*{1.50em}- Use `coupled\_with: [unit\_id]` when this unit's validity depends on the STATUS of another unit (active/closed/superseded). \\
\hspace*{2.50em}Example: "Netflix autopay on Card 3" is coupled\_with the unit declaring Card 3's status — if Card 3 closes, the Netflix routing must be re-evaluated. \\
\hspace*{1.50em}- Use `derived\_from: [unit\_id]` when this unit's VALUE was computed from another unit's value. \\
\hspace*{2.50em}Example: "monthly savings \$1,800" is derived\_from the unit "take-home pay \$6,000" via a savings-rate rule — if take-home changes, savings derives differently. \\
\hspace*{1.50em}- Use `triggers: [\{"kind": "date\_passed", "payload": \{"date": "YYYY-MM-DD"\}\}]` when this unit becomes operationally relevant on a specific date or condition. \\
\hspace*{2.50em}Allowed kinds: "date\_passed", "entity\_closed", "supersession\_announced", "cascade".

5. NO CHANGE: If the turn contains no new state-relevant information, or if the state already captures everything.

\#\# Rules: \\
- Only create units for information RELEVANT to future decisions \\
- Do NOT duplicate information already in an existing unit — check the current state first \\
- Priority "hard" for: verified facts, binding commitments, remaining obligations, things that constrain future actions \\
- Priority "soft" for: preferences, nice-to-haves, things that can yield \\
- Keep content concise — one clear sentence per unit \\
- Type and scope are free-form labels you assign \\
- For coupled\_with / derived\_from / triggers: only include when there is a REAL dependency. Leave empty otherwise.

\#\# Output format: \\
Return a JSON object: \\
\{ \\
\hspace*{1.00em}"add": [ \\
\hspace*{2.00em}\{ \\
\hspace*{3.00em}"content": "string", \\
\hspace*{3.00em}"priority": "hard" | "soft", \\
\hspace*{3.00em}"source\_type": "user" | "action" | "tool\_result", \\
\hspace*{3.00em}"type": "string (free-form category label)", \\
\hspace*{3.00em}"scope": "string (what this applies to)", \\
\hspace*{3.00em}"coupled\_with": ["unit\_id\_a", "unit\_id\_b"], \\
\hspace*{3.00em}"derived\_from": ["unit\_id\_c"], \\
\hspace*{3.00em}"triggers": [\{"kind": "date\_passed", "payload": \{"date": "2026-06-15"\}, "description": "tax deadline"\}] \\
\hspace*{2.00em}\} \\
\hspace*{1.00em}], \\
\hspace*{1.00em}"supersede": [ \\
\hspace*{2.00em}\{ \\
\hspace*{3.00em}"unit\_id": "string (ID of existing unit to supersede)", \\
\hspace*{3.00em}"reason": "string (why it's being superseded)", \\
\hspace*{3.00em}"replacement": \{ \\
\hspace*{4.00em}"content": "string (new version)", \\
\hspace*{4.00em}"priority": "hard" | "soft", \\
\hspace*{4.00em}"source\_type": "user" | "action" | "tool\_result", \\
\hspace*{4.00em}"type": "string", \\
\hspace*{4.00em}"scope": "string", \\
\hspace*{4.00em}"coupled\_with": [], \\
\hspace*{4.00em}"derived\_from": [], \\
\hspace*{4.00em}"triggers": [] \\
\hspace*{3.00em}\} \\
\hspace*{2.00em}\} \\
\hspace*{1.00em}] \\
\}

If no updates are needed, return: \{"add": [], "supersede": []\} \\
Output ONLY the JSON object, no other text.
\end{tcolorbox}

\subsection{PolicyEncoder}
\label{app:policy-encoder-prompt}

The PolicyEncoder runs once per scenario over the policy document (if any) to extract binding rules into the same \texttt{StateUnit} schema as the TurnEncoder. It plugs in the same \textbf{Task Context} block as above. Note that of the evaluated benchmarks, only Tau2 \citep{barres2025tau} actually has specific policy (which should be treated differently from typical evolving state).

\begin{tcolorbox}[smbprompt, title=PolicyEncoder System Prompt]
\ttfamily\small
\setlength{\parskip}{4pt}
\setlength{\parindent}{0pt}
You are a constraint extraction system. Given a policy document for a task-executing agent, extract every rule, constraint, and requirement as a structured unit.

For each unit, determine: \\
- content: the rule stated clearly and completely in one sentence \\
- priority: "hard" if violating this would make the task fail or break policy, "soft" if it is a preference or guideline that can yield under pressure \\
- type: a short free-form label describing the category (e.g., "verification", "payment", "eligibility", "cancellation", "communication") \\
- scope: what this rule applies to (e.g., "all\_reservations", "basic\_economy", "gold\_members", "cancellation\_requests")

Rules for priority assignment: \\
- "hard": explicit prohibitions ("must not", "cannot", "only if"), eligibility checks, verification requirements, safety rules, actions the agent must or must not take \\
- "soft": guidelines about tone, order of operations when flexible, optional offers, preferences

\{task\_context\}

\#\# Policy Document: \\
\{policy\}

\#\# Output format: \\
Return a JSON array of objects. Each object has: \\
\hspace*{1.00em}"content": string, \\
\hspace*{1.00em}"priority": "hard" | "soft", \\
\hspace*{1.00em}"type": string, \\
\hspace*{1.00em}"scope": string

Extract ALL rules. Do not summarize or merge rules — each distinct constraint gets its own unit. If a paragraph contains multiple rules, split them.

Output ONLY the JSON array, no other text.
\end{tcolorbox}

\subsection{State Injection (at probe time)}
\label{app:state-injection-prompts}

At probe time the agent's system message is assembled with an optional per-scenario \texttt{system\_prompt} (only used if there is per-scenario policy to encode), followed by the rendered \textbf{State Block}, followed by the \textbf{Recompute Guidance}. The state block is produced by a deterministic renderer (no LLM); its layout is shown below.

\begin{tcolorbox}[smbprompt, title=State Injection --- Block Layout]
\ttfamily\small
\setlength{\parskip}{4pt}
\setlength{\parindent}{0pt}
\#\# BINDING — THIS CONVERSATION \\
- [<type>] <unit content> \\
- [<type>] <unit content> \\
...

\#\# BINDING — POLICY RULES \\
- [<type>] <unit content> \\
- [<type>] <unit content> \\
...

\#\# PREFERENCES (yield if necessary) \\
- [<type>] <unit content> \\
...

\#\# ! NEEDS RECHECK — may be stale due to recent events \\
- [<type>] <unit content> \\
\hspace*{1.00em}-> trigger: <reason> \\
- [<type>] <unit content> \\
\hspace*{1.00em}-> trigger: <reason> \\
...
\end{tcolorbox}

\begin{tcolorbox}[smbprompt, title=State Injection --- Recompute Guidance]
\ttfamily\small
\setlength{\parskip}{4pt}
\setlength{\parindent}{0pt}
When answering, work through each relevant piece of state. Specifically: \\
\hspace*{0.50em}- Hard constraints (medical, scope-bound) take priority over recent soft preferences. \\
\hspace*{0.50em}- A unit flagged NEEDS RECHECK is STALE: an input it was derived \\
\hspace*{1.50em}from has changed, so its baked-in value is now wrong. Do NOT use \\
\hspace*{1.50em}that number. RECOMPUTE it: take its stated derivation (e.g. \\
\hspace*{1.50em}'X = A * 2'), substitute the CURRENT value of each input from the \\
\hspace*{1.50em}active state above, and evaluate. If the result feeds another \\
\hspace*{1.50em}NEEDS RECHECK unit, recompute that one from the value you just \\
\hspace*{1.50em}derived — propagate through the chain in order before answering. \\
\hspace*{1.50em}Only recompute units on the path to the question; leave others. \\
\hspace*{0.50em}- For cascade-coupled units, surface the downstream implication explicitly.
\end{tcolorbox}

\section{StateMemWrapper}
\label{app:wrapper}

StateMemWrapper carries \textsc{StateMem}'s three resolution
primitives --- supersession precedence, rule-over-instance ranking, and
derived-value recomputation --- into a single answer-time prompt. State
is resolved lazily, scoped to the question, from the turn-numbered
transcript: the chronology that decides which of two conflicting values
is operative survives because the value chain is extracted \emph{with}
its turn ordering, at the moment it is needed. This yields the property
the composition sweep exploits: the full StateMem treatment attaches to
any backend as a rewrite of the one answer call the backend already
makes --- no store, no per-turn processing, no added LLM calls, and a
call count identical to the backend alone.

\subsection{Mechanism}
\label{app:wrapper-mechanism}\label{app:wrapper-prompts}
The answer call works in two committed sections. Section~1
(\emph{trace}, capped at $250$ words) reconstructs, in chronological
order, every turn that establishes, updates, or supersedes the entities
the question touches, each cited as \texttt{[turn N]}; it must include
the standing rules that govern the decision and the most recent stated
value of every input those rules apply to, note derived values whose
inputs later changed, and end with the current operative value of each
relevant entity. The trace is committed before any answer is produced,
which forecloses post-hoc rationalization: the model binds itself to a
state reading first and answers from it second. Section~2
(\emph{resolve}) commits the answer under four precedence rules ---
later supersedes earlier; standing rules outrank past instances,
applied to current inputs; derived values are recomputed, never quoted;
a fact is retired only by explicit supersession or expiry, and emits
one final \texttt{ANSWER:} line, which is parsed as the prediction. The
full prompt:

\begin{tcolorbox}[smbprompt, title=StateMemWrapper --- Fused Answer Prompt (Trace $+$ Resolve)]
\ttfamily\small
\setlength{\parskip}{4pt}
\setlength{\parindent}{0pt}
\{system\_prompt\}

You are answering a question about a long conversation using state tracing. \\
\#\# Question: \{question\} \\
\#\# Conversation transcript: \{transcript\}

Work in TWO sections, in order:

\#\# Section 1 --- State trace (under 250 words): \\
List, in chronological order, every turn that establishes, updates, or
supersedes the entities the question asks about ([turn N] speaker: what
changed). Include standing rules that govern the decision AND the most
recent stated value of every input those rules apply to (amounts,
quantities, dates, thresholds) --- even if mentioned only once or in
passing; scan for them before concluding an input is unknown. Note
derived values whose inputs later changed. End with the current
operative value of each relevant entity. Commit to the trace BEFORE
answering.

\#\# Section 2 --- Resolution and answer: \\
Apply, in order: \\
\hspace*{0.50em}(1) later supersedes earlier; \\
\hspace*{0.50em}(2) standing rules outrank past one-off instances --- apply \\
\hspace*{1.50em}the rule to CURRENT inputs; \\
\hspace*{0.50em}(3) recompute derived values from current inputs, never \\
\hspace*{1.50em}reuse stale cached numbers; \\
\hspace*{0.50em}(4) a fact is only retired if actually superseded or expired.

End with exactly one final line: \\
ANSWER: \textless the specific value or decision the question asks for\textgreater
\end{tcolorbox}

\begin{tcolorbox}[smbprompt, title=Wrapper-Ctrl --- Matched Control (structure removed)]
\ttfamily\small
\setlength{\parskip}{4pt}
\setlength{\parindent}{0pt}
\{system\_prompt\}

You are answering a question about a long conversation. \\
\#\# Question: \{question\} \\
\#\# Conversation transcript: \{transcript\}

Work in TWO sections, in order:

\#\# Section 1 --- Relevant information (under 250 words): \\
Summarize the information in the conversation that is relevant to the
question. Commit to this summary BEFORE answering.

\#\# Section 2 --- Answer: \\
End with exactly one final line: \\
ANSWER: \textless the specific value or decision the question asks for\textgreater
\end{tcolorbox}

\subsection{Control}
\label{app:wrapper-ctrl}
\textsc{wrapper-ctrl} is matched on every surface except structure:
the same system prompt, the same two-section format, the same $250$-word
Section~1 budget, the same commit-before-answering instruction, the same
final \texttt{ANSWER:} line, and the same transcript-plus-chunks
context. Its Section~1 asks only to ``summarize the information in the
conversation that is relevant to the question'', no turn numbers, no
value chains, no operative-state section, and its Section~2 contains
no precedence rules. The wrapper$-$control delta therefore isolates
state structure at equal call count, token budget, and context.

\subsection{Composition protocol}
\label{app:wrapper-protocol}
A backend attaches through a two-method interface:
\begin{quote}\footnotesize\ttfamily
class MemoryBackend(Protocol):\\
\hspace*{1em}def ingest(self, text) -> None: ...\\
\hspace*{1em}def search(self, query, k=10) -> list[str]: ...
\end{quote}
The backend ingests every turn with its unmodified native pipeline; at
the probe, its top-$k$ chunks are appended to the transcript under a
framing header matched to the backend type (retrieval excerpts for
BM25/Dense, extracted memories for Mem0/A-Mem/LightMem/MemoryOS). The
wrapper consumes only strings, so lexical, dense, and LLM-extracted
backends satisfy the identical interface, and switching backends is a
constructor swap. The transcript is rendered as numbered turns with a
$400$k-character tail guard so the longest histories fit both
backbones' context windows.

\subsection{Sweep protocol}
\label{app:wrapper-sweep}
Each backend ingests each conversation once; all three answer
conditions (alone, +Ctrl, +SMW) are then issued against the same
populated store, so condition deltas cannot arise from ingest variance.
Case sets are frozen $n{=}60$ manifests per benchmark
(\textsc{StateMemBench}: $30$ Set-A probes plus $30$ Set-B probes
sampled from $22$ long scenarios---$16$ contributing one thread-probe,
$4$ two, $2$ all three---at $k{=}10$; LongMemEval: multi-session
question type, $k{=}20$), identical across backends, conditions, and
backbones. The frozen set's category mix differs from the full
benchmark's (anti-trap is overweighted, status underweighted), so
absolute scores in the sweep are mix-dependent and sit above the
full-benchmark level; the within-row condition deltas are the quantity
of interest. A single fixed deepseek-v4-pro judge scores all conditions
in one pass, and every wrapper output is logged with its full state
trace, so any scored answer can be audited against the trace that
produced it.

\subsection{Cost}
  \label{sec:wrapper-cost}
  The wrapper adds no LLM calls and no ingest-time work: it rewrites the
  single answer call the backend already makes, so call count, storage,
  and per-turn cost are unchanged, and overhead is confined to that one
  call. There it consists of two bounded constants: a fixed instruction
  block of $155$ input tokens --- measured constant across both
  benchmarks and all six backends ($360$ answers per condition per
  benchmark) --- and a trace capped at $250$ words, observed at
  $169$--$188$ output tokens beyond the matched control. The full state
  treatment therefore costs ${\sim}350$ tokens per question, independent
  of conversation length, backend, and benchmark: on LongMemEval this is
  $0.2\%$ of the answer prompt, purchasing the condition deltas of
  Table~\ref{tab:wrapper}.

  The wrapper does condition on the transcript, so its answer call is
  transcript-sized (median input $8.6$k tokens on \textsc{StateMemBench},
  $87$k on LongMemEval) where a backend-alone call sees only retrieved
  chunks. This is the cost of the long-context baseline, not of the
  wrapper: a long-context answer pays the same input, and the wrapper
  matches it within $155$ tokens while adding the backend's retrieval and
  the state treatment. The matched control makes the separation exact, 
  control and wrapper differ by $155$ input and ${\sim}180$ output tokens,
  so every accuracy delta between them in Table~\ref{tab:wrapper} is
  bought at that price. Median total output, trace and answer included,
  is $398$--$416$ tokens.

\section{Results}

\begin{table*}[!t]
\centering
\small
\setlength{\tabcolsep}{3pt}
\renewcommand{\arraystretch}{1.12}
\caption{State-agreement accuracy on \textsc{StateMemBench} with deepseek-v4-pro judge.
Left block: accuracy per horizon and $n$-weighted overall. Middle block:
accuracy by probe category over all 322 probes (status $n{=}116$, sequence
$n{=}72$, salience $n{=}51$, anti-trap $n{=}39$, compound $n{=}44$); Right block: accuracy by domain
(finance $n{=}93$, shopping $n{=}127$, research $n{=}102$). Dense retrieval
embeds with OpenAI text-embedding-3-small on both substrates; answers come from the panel's
substrate model. Per column within each panel, best in \textbf{bold},
second best \underline{underlined}; StateMem ablations excluded
from marking.}
\label{tab:state-mem-bench-full}
\begin{tabular}{c l ccc ccccc ccc}
\toprule
& & \multicolumn{3}{c}{\textbf{By horizon}} & \multicolumn{5}{c}{\textbf{By category}} & \multicolumn{3}{c}{\textbf{By domain}} \\
\cmidrule(lr){3-5}\cmidrule(lr){6-10}\cmidrule(lr){11-13}
& \textbf{Condition} & Short & Long & Overall & Status & Sequence & Salience & Anti-trap & Compound & Finance & Shopping & Research \\
\midrule
\multirow{4}{*}{\rlab{Long-context}}
 & Qwen-3.5-9B        & \underline{0.137} & 0.167 & \underline{0.149} & 0.078 & 0.000 & \underline{0.098} & \textbf{0.692} & \textbf{0.159} & 0.118 & 0.197 & \textbf{0.118} \\
 & Qwen-3.6-35B-A3B   & 0.063 & 0.146 & 0.097 & 0.026 & 0.000 & 0.059 & 0.632 & 0.023 & 0.110 & 0.165 & 0.000 \\
 & GPT-5.4-Nano       & \textbf{0.311} & \textbf{0.227} & \textbf{0.277} & \textbf{0.172} & \textbf{0.153} & \textbf{0.569} & 0.615 & \underline{0.114} & \textbf{0.344} & \textbf{0.402} & \underline{0.059} \\
 & DeepSeek-V4-Flash  & 0.132 & \underline{0.174} & \underline{0.149} & \underline{0.103} & \underline{0.056} & 0.059 & \underline{0.641} & 0.091 & \underline{0.161} & \underline{0.244} & 0.020 \\
\midrule
\multirow{16}{*}{\rlab{Qwen-3.5-9B}}
 & No memory          & 0.000 & 0.015 & 0.006 & 0.009 & 0.000 & 0.000 & 0.026 & 0.000 & 0.000 & 0.000 & 0.020 \\
\cmidrule(l){2-13}
 & BM25               & 0.095 & 0.167 & 0.125 & 0.034 & 0.042 & 0.000 & 0.820 & 0.023 & 0.107 & 0.102 & 0.167 \\
 & Dense retrieval    & 0.105 & 0.167 & 0.130 & 0.034 & 0.014 & 0.000 & \textbf{0.923} & 0.023 & 0.118 & 0.126 & 0.147 \\
\cmidrule(l){2-13}
 & nano-graphrag      & 0.126 & 0.167 & 0.143 & 0.035 & 0.000 & \underline{0.100} & 0.895 & 0.068 & 0.140 & 0.149 & 0.137 \\
 & HippoRAG           & 0.105 & 0.167 & 0.130 & 0.034 & 0.014 & 0.000 & 0.795 & \textbf{0.136} & 0.118 & 0.126 & 0.147 \\
 & LightRAG           & 0.089 & 0.152 & 0.115 & 0.026 & 0.014 & 0.059 & 0.769 & 0.000 & 0.065 & 0.126 & 0.147 \\
 & GraphRAG           & \underline{0.216} & \textbf{0.235} & \underline{0.224} & \underline{0.164} & \underline{0.069} & \textbf{0.118} & \textbf{0.923} & \textbf{0.136} & \underline{0.204} & \underline{0.197} & \textbf{0.275} \\
\cmidrule(l){2-13}
 & Mem0               & 0.116 & 0.197 & 0.149 & 0.026 & 0.042 & 0.039 & \underline{0.897} & \underline{0.114} & 0.151 & 0.126 & 0.177 \\
 & A-Mem              & 0.100 & 0.167 & 0.127 & 0.008 & 0.042 & 0.000 & \underline{0.897} & 0.045 & 0.118 & 0.138 & 0.138 \\
 & LightMem           & 0.021 & 0.015 & 0.019 & 0.017 & 0.000 & 0.000 & 0.102 & 0.000 & 0.000 & 0.016 & 0.039 \\
 & MemoryOS           & 0.026 & 0.023 & 0.025 & 0.025 & 0.000 & 0.000 & 0.102 & 0.023 & 0.022 & 0.008 & 0.049 \\
\hcrow
 & \textbf{StateMem (ours)} & \textbf{0.237} & \underline{0.227} & \textbf{0.233} & \textbf{0.181} & \textbf{0.208} & 0.020 & 0.846 & \underline{0.114} & \textbf{0.269} & \textbf{0.205} & \underline{0.235} \\
 & \abl{-- extraction-only}      & 0.126 & 0.197 & 0.155 & 0.086 & 0.000 & 0.078 & 0.795 & 0.114 & 0.161 & 0.126 & 0.186 \\
 & \abl{-- + supersession}       & 0.137 & 0.220 & 0.171 & 0.112 & 0.028 & 0.039 & 0.872 & 0.091 & 0.172 & 0.165 & 0.176 \\
 & \abl{-- w/o dep.-prop.}       & 0.202 & 0.250 & 0.222 & 0.155 & 0.139 & 0.040 & 0.897 & 0.140 & 0.215 & 0.238 & 0.207 \\
 & \abl{-- w/o recompute}        & 0.195 & 0.227 & 0.208 & 0.190 & 0.097 & 0.059 & 0.846 & 0.045 & 0.204 & 0.213 & 0.206 \\
\midrule
\multirow{16}{*}{\rlab{deepseek-v4-flash}}
 & No memory          & 0.000 & 0.008 & 0.003 & 0.000 & 0.000 & 0.000 & 0.026 & 0.000 & 0.000 & 0.000 & 0.010 \\
\cmidrule(l){2-13}
 & BM25               & 0.105 & 0.197 & 0.143 & 0.034 & 0.097 & 0.000 & 0.872 & 0.023 & 0.108 & 0.110 & 0.216 \\
 & Dense retrieval    & 0.168 & \underline{0.258} & \underline{0.205} & 0.086 & 0.181 & 0.000 & \textbf{1.000} & 0.091 & 0.194 & \underline{0.228} & 0.186 \\
\cmidrule(l){2-13}
 & nano-graphrag      & \underline{0.174} & 0.197 & 0.183 & 0.095 & 0.014 & \underline{0.098} & \underline{0.949} & \underline{0.114} & \underline{0.215} & 0.173 & 0.167 \\
 & HippoRAG           & 0.132 & \underline{0.258} & 0.184 & 0.086 & 0.139 & 0.000 & 0.923 & 0.068 & 0.151 & 0.173 & 0.225 \\
 & LightRAG           & 0.058 & 0.144 & 0.093 & 0.026 & 0.028 & 0.020 & 0.615 & 0.000 & 0.065 & 0.126 & 0.078 \\
 & GraphRAG           & 0.137 & 0.227 & 0.174 & \underline{0.103} & 0.042 & \textbf{0.118} & 0.692 & \textbf{0.182} & 0.097 & 0.126 & \underline{0.304} \\
\cmidrule(l){2-13}
 & Mem0               & 0.168 & 0.189 & 0.177 & \underline{0.103} & 0.028 & \underline{0.098} & \underline{0.949} & 0.023 & 0.151 & 0.205 & 0.167 \\
 & A-Mem              & 0.158 & \underline{0.258} & 0.199 & 0.060 & \underline{0.264} & 0.000 & \underline{0.949} & 0.023 & 0.194 & 0.197 & 0.206 \\
 & LightMem           & 0.021 & 0.000 & 0.012 & 0.017 & 0.000 & 0.000 & 0.051 & 0.000 & 0.000 & 0.016 & 0.020 \\
 & MemoryOS           & 0.011 & 0.045 & 0.025 & 0.009 & 0.000 & 0.000 & 0.154 & 0.023 & 0.011 & 0.039 & 0.020 \\
\hcrow
 & \textbf{StateMem (ours)} & \textbf{0.411} & \textbf{0.295} & \textbf{0.363} & \textbf{0.302} & \textbf{0.347} & \textbf{0.216} & 0.872 & \textbf{0.273} & \textbf{0.366} & \textbf{0.339} & \textbf{0.392} \\
 & \abl{-- extraction-only}      & 0.153 & 0.205 & 0.174 & 0.078 & 0.028 & 0.157 & 0.744 & 0.182 & 0.215 & 0.157 & 0.157 \\
 & \abl{-- + supersession}       & 0.337 & 0.242 & 0.298 & 0.241 & 0.250 & 0.157 & 0.846 & 0.205 & 0.280 & 0.283 & 0.333 \\
 & \abl{-- w/o dep.-prop.}       & 0.400 & 0.333 & 0.373 & 0.336 & 0.375 & 0.196 & 0.846 & 0.250 & 0.333 & 0.362 & 0.422 \\
 & \abl{-- w/o recompute}        & 0.326 & 0.265 & 0.301 & 0.267 & 0.222 & 0.196 & 0.897 & 0.114 & 0.301 & 0.244 & 0.407 \\
\bottomrule
\end{tabular}
\end{table*}

\begin{table*}[!t]
\centering
\small
\setlength{\tabcolsep}{4.5pt}
\renewcommand{\arraystretch}{1.12}
\caption{Accuracy by question type on LongMemEval (top; full set, $n{=}500$,
$k{=}20$) and LoCoMo (bottom; $n{=}1{,}985$; all retrieval $k{=}20$), both
substrates (thinking off), graded by a fixed deepseek-v4-pro judge.
Question-type $n$ in the header. Knowledge-update and temporal-reasoning
(LME) and temporal-reasoning (LoCoMo) are the drift-shaped subsets; LoCoMo
adversarial rewards abstention. deepseek LME memory-system cells are from the
final unified $n{=}500$ judge pass (all arms scored together).
Per column within each panel, best in \textbf{bold}, second best
\underline{underlined}.}
\label{tab:lme-locomo-qtype}

\begin{tabular}{c l c cccccc}
\toprule
& & & \multicolumn{6}{c}{\textbf{LongMemEval question type}} \\
\cmidrule(lr){4-9}
& \textbf{Condition} & \textbf{Overall} & \shortstack{SS-user\\(70)} & \shortstack{SS-asst.\\(56)} & \shortstack{SS-pref.\\(30)} & \shortstack{Multi-sess.\\(133)} & \shortstack{Know.-upd.\\(78)} & \shortstack{Temporal\\(133)} \\
\midrule
\multirow{7}{*}{\rlab{Qwen-3.5-9B}}
 & Long context & 0.550 & \textbf{0.900} & \textbf{0.946} & 0.333 & 0.436 & \textbf{0.744} & 0.248 \\
 & No memory    & 0.062 & 0.086 & 0.054 & 0.067 & 0.053 & 0.038 & 0.075 \\
 & Mem0         & \underline{0.566} & 0.871 & 0.607 & \underline{0.533} & \textbf{0.602} & 0.603 & \underline{0.338} \\
 & A-Mem        & 0.520 & \underline{0.886} & \underline{0.929} & 0.467 & 0.361 & 0.679 & 0.233 \\
 & LightMem     & 0.064 & 0.086 & 0.073 & 0.000 & 0.053 & 0.038 & 0.092 \\
 & MemoryOS     & 0.080 & 0.114 & 0.036 & 0.100 & 0.060 & 0.154 & 0.053 \\
\hcrow
 & \textbf{StateMem (ours)} & \textbf{0.580} & 0.871 & 0.536 & \textbf{0.700} & \underline{0.519} & \underline{0.718} & \textbf{0.398} \\
\midrule
\multirow{7}{*}{\rlab{deepseek-flash}}
 & Long context & \textbf{0.666} & \textbf{0.957} & \textbf{1.000} & \textbf{0.600} & \underline{0.594} & \underline{0.782} & \underline{0.391} \\
 & No memory    & 0.056 & 0.086 & 0.018 & 0.000 & 0.083 & 0.038 & 0.053 \\
 & Mem0         & 0.594 & 0.929 & 0.589 & 0.367 & \textbf{0.617} & 0.692 & \underline{0.391} \\
 & A-Mem        & 0.510 & 0.886 & \underline{0.875} & 0.333 & 0.391 & 0.679 & 0.218 \\
 & LightMem     & 0.056 & 0.086 & 0.018 & 0.000 & 0.083 & 0.038 & 0.053 \\
 & MemoryOS     & 0.072 & 0.114 & 0.000 & 0.000 & 0.075 & 0.179 & 0.030 \\
\hcrow
 & \textbf{StateMem (ours)} & \underline{0.656} & \underline{0.943} & 0.607 & \underline{0.400} & 0.534 & \textbf{0.795} & \textbf{0.624} \\
\bottomrule
\end{tabular}

\vspace{5pt}

\begin{tabular}{c l c ccccc}
\toprule
& & & \multicolumn{5}{c}{\textbf{LoCoMo question type}} \\
\cmidrule(lr){4-8}
& \textbf{Condition} & \textbf{Overall} & \shortstack{Single-hop\\(840)} & \shortstack{Multi-hop\\(282)} & \shortstack{Temporal\\(321)} & \shortstack{Open-dom.\\(96)} & \shortstack{Adversarial\\(446)} \\
\midrule
\multirow{7}{*}{\rlab{Qwen-3.5-9B}}
 & Long context & \textbf{0.612} & \textbf{0.865} & \underline{0.445} & \underline{0.412} & \underline{0.438} & \underline{0.422} \\
 & No memory    & 0.010 & 0.008 & 0.011 & 0.000 & 0.042 & 0.013 \\
 & Mem0         & 0.481 & 0.637 & 0.376 & \textbf{0.595} & \textbf{0.448} & 0.182 \\
 & A-Mem        & 0.266 & 0.355 & 0.131 & 0.025 & 0.281 & 0.357 \\
 & LightMem     & 0.027 & 0.021 & 0.039 & 0.006 & 0.115 & 0.027 \\
 & MemoryOS     & 0.040 & 0.042 & 0.050 & 0.009 & 0.146 & 0.029 \\
\hcrow
 & \textbf{StateMem (ours)} & \underline{0.566} & \underline{0.790} & \textbf{0.461} & 0.296 & 0.385 & \textbf{0.442} \\
\midrule
\multirow{7}{*}{\rlab{deepseek-flash}}
 & Long context & \underline{0.587} & \textbf{0.870} & \textbf{0.482} & \textbf{0.371} & \underline{0.479} & 0.298 \\
 & No memory    & 0.005 & 0.001 & 0.004 & 0.000 & 0.031 & 0.009 \\
 & Mem0         & 0.462 & 0.690 & 0.401 & 0.336 & 0.458 & 0.161 \\
 & A-Mem        & 0.279 & 0.379 & 0.131 & 0.028 & 0.219 & \textbf{0.377} \\
 & LightMem     & 0.020 & 0.014 & 0.032 & 0.012 & 0.073 & 0.018 \\
 & MemoryOS     & 0.031 & 0.026 & 0.050 & 0.009 & 0.146 & 0.018 \\
\hcrow
 & \textbf{StateMem (ours)} & \textbf{0.592} & \underline{0.854} & \underline{0.473} & \textbf{0.371} & \textbf{0.531} & \underline{0.354} \\
\bottomrule
\end{tabular}
\end{table*}

\begin{table*}[!t]
\centering
\small
\setlength{\tabcolsep}{5pt}
\renewcommand{\arraystretch}{1.05}
\caption{\textbf{StateMemWrapper sweep on \textsc{StateMemBench}, by probe category and
domain} (gold \%, frozen $n{=}60$ mixed set; category $n$: status 12, sequence 12, salience 13, anti-trap 13,
compound 10; domain $n$: finance 22, shopping 18, research 20; deepseek-v4-pro judge). Anti-trap probes are
controls (correct behaviour is \emph{not} updating), hence high
accuracy even for weak arms — and +SMW does \emph{not} inflate there,
i.e.\ it updates selectively rather than aggressively. Per-cell $n$ is
$10$--$13$; splits are exploratory. +SMW rows shaded.}
\label{tab:wrapper-split-ourbench}
\begin{tabular}{c l l ccccc ccc}
\toprule
& & & \multicolumn{5}{c}{\textbf{By category}} & \multicolumn{3}{c}{\textbf{By domain}} \\
\cmidrule(lr){4-8}\cmidrule(lr){9-11}
& \textbf{Backend} & \textbf{Cond.} & Status & Seq. & Sal. & Anti & Comp. & Fin. & Shop. & Res. \\
\midrule
\multirow{18}{*}{\rlab{Qwen-3.5-9B}}
 & \multirow{3}{*}{Mem0}     & alone &  0 &  8 &  0 &  92 & 10 & 27 & 17 & 20 \\
 &                           & +Ctrl & 17 & 17 & 23 &  92 &  0 & 23 & 33 & 40 \\
\hcrow & & +SMW                      & 50 & 58 & 77 &  92 & 40 & 64 & 56 & 70 \\
 & \multirow{3}{*}{A-Mem}    & alone &  0 &  0 &  0 & 100 & 10 & 18 & 28 & 20 \\
 &                           & +Ctrl &  8 & 25 & 23 & 100 & 10 & 27 & 28 & 50 \\
\hcrow & & +SMW                      & 25 & 50 & 62 &  85 & 60 & 55 & 44 & 70 \\
 & \multirow{3}{*}{LightMem} & alone &  0 &  0 &  0 &  15 &  0 &  0 &  0 &  0 \\
 &                           & +Ctrl &  8 & 33 & 15 & 100 &  0 & 23 & 22 & 60 \\
\hcrow & & +SMW                      & 42 & 58 & 46 &  85 & 40 & 50 & 56 & 70 \\
 & \multirow{3}{*}{MemOS}    & alone &  0 &  0 &  0 &  23 &  0 &  5 &  0 & 10 \\
 &                           & +Ctrl &  0 &  8 & 31 &  77 &  0 & 23 & 22 & 30 \\
\hcrow & & +SMW                      & 42 & 50 & 54 &  92 & 40 & 55 & 56 & 60 \\
 & \multirow{3}{*}{BM25}     & alone &  0 &  0 &  0 &  92 &  0 & 18 & 17 & 20 \\
 &                           & +Ctrl &  0 & 25 & 31 &  85 & 10 & 23 & 28 & 50 \\
\hcrow & & +SMW                      & 42 & 58 & 54 &  92 & 30 & 50 & 50 & 70 \\
 & \multirow{3}{*}{Dense}    & alone &  0 &  0 &  0 &  23 &  0 &  0 &  0 & 10 \\
 &                           & +Ctrl & 17 & 25 & 31 &  85 & 10 & 27 & 39 & 50 \\
\hcrow & & +SMW                      & 33 & 50 & 46 &  85 & 40 & 50 & 50 & 60 \\
\midrule
\multirow{18}{*}{\rlab{deepseek-v4-flash}}
 & \multirow{3}{*}{Mem0}     & alone & 17 &  8 &  0 &  92 & 20 & 32 & 28 & 20 \\
 &                           & +Ctrl & 17 & 50 & 69 &  85 & 60 & 64 & 50 & 50 \\
\hcrow & & +SMW                      & 42 & 67 & 77 &  92 & 80 & 73 & 67 & 90 \\
 & \multirow{3}{*}{A-Mem}    & alone & 25 & 17 &  0 & 100 & 20 & 36 & 33 & 20 \\
 &                           & +Ctrl & 17 & 33 & 46 &  92 & 70 & 55 & 50 & 50 \\
\hcrow & & +SMW                      & 42 & 67 & 92 &  85 & 90 & 86 & 67 & 80 \\
 & \multirow{3}{*}{LightMem} & alone &  0 &  0 &  0 &   8 &  0 &  0 &  0 & 10 \\
 &                           & +Ctrl &  8 & 25 & 46 &  77 & 70 & 55 & 44 & 50 \\
\hcrow & & +SMW                      & 50 & 67 & 77 &  92 & 80 & 82 & 56 & 80 \\
 & \multirow{3}{*}{MemOS}    & alone &  0 &  0 &  0 &   8 &  0 &  5 &  0 &  0 \\
 &                           & +Ctrl & 25 & 25 & 54 &  85 & 60 & 55 & 50 & 40 \\
\hcrow & & +SMW                      & 58 & 58 & 62 &  92 & 70 & 68 & 56 & 90 \\
 & \multirow{3}{*}{BM25}     & alone &  0 &  0 &  0 &  92 & 10 & 18 & 22 & 20 \\
 &                           & +Ctrl & 25 & 33 & 46 &  85 & 50 & 59 & 33 & 50 \\
\hcrow & & +SMW                      & 58 & 50 & 77 &  85 & 80 & 77 & 67 & 80 \\
 & \multirow{3}{*}{Dense}    & alone &  0 &  0 &  0 &   0 &  0 &  0 &  0 &  0 \\
 &                           & +Ctrl &  8 & 33 & 46 &  85 & 60 & 50 & 39 & 50 \\
\hcrow & & +SMW                      & 33 & 58 & 69 &  85 & 70 & 68 & 56 & 90 \\
\bottomrule
\end{tabular}
\end{table*}

\begin{table}[!t]
\centering
\small
\setlength{\tabcolsep}{4pt}
\renewcommand{\arraystretch}{1.08}
\caption{\textbf{StateMemWrapper sweep on LongMemEval: paired outcome
transitions vs.\ the backend alone} (frozen $n{=}60$; Dense/Qwen has
$n{=}59$, one case failed at ingest. The set is drawn from
the multi-session question type by construction, so no category split
applies). For each backend, each condition's answers are compared case-by-case
against the \emph{alone} answers: Fix = alone wrong $\to$ condition right;
Brk = alone right $\to$ condition wrong; Fix $-$ Brk equals the net case
change in Table~\ref{tab:wrapper}. +SMW is net-positive for every
backend on both substrates; the control is net-negative for three Qwen
backends (Mem0, A-Mem, Dense).}
\label{tab:wrapper-lme-transitions}
\begin{tabular}{l cc cc c cc cc}
\toprule
& \multicolumn{4}{c}{\textbf{Qwen-3.5-9B}} & & \multicolumn{4}{c}{\textbf{deepseek-v4-flash}} \\
\cmidrule(lr){2-5}\cmidrule(lr){7-10}
& \multicolumn{2}{c}{+Ctrl} & \multicolumn{2}{c}{+SMW} & & \multicolumn{2}{c}{+Ctrl} & \multicolumn{2}{c}{+SMW} \\
\cmidrule(lr){2-3}\cmidrule(lr){4-5}\cmidrule(lr){7-8}\cmidrule(lr){9-10}
\textbf{Backend} & Fix$\uparrow$ & Brk$\downarrow$ & Fix$\uparrow$ & Brk$\downarrow$ & & Fix$\uparrow$ & Brk$\downarrow$ & Fix$\uparrow$ & Brk$\downarrow$ \\
\midrule
Mem0     & 10 & 11 & \textbf{18} & 12 & & 15 & 7 & 15 & \textbf{6} \\
A-Mem    &  7 & 13 & \textbf{17} & \textbf{8} & & \textbf{18} & 8 & 16 & \textbf{4} \\
LightMem &  4 &  0 & \textbf{14} & \textbf{0} & & 17 & \textbf{0} & \textbf{20} & 1 \\
MemOS    &  4 &  3 & \textbf{10} & \textbf{0} & & 17 & 2 & \textbf{20} & 2 \\
BM25     &  7 &  7 & \textbf{15} & \textbf{3} & & 16 & 2 & \textbf{17} & 2 \\
Dense    &  8 & 17 & \textbf{12} & \textbf{10} & & 12 & 7 & \textbf{15} & 13 \\
\bottomrule
\end{tabular}
\end{table}
\subsection{Evaluated Baselines}
\label{app:evaluation}
All baselines call the same answer template as the no-memory floor. Whenever we retrieve, we set $k=10$ for all baselines that take $k$ for \textsc{StateMemBench} and set $k=20$ for the relatively longer scenarios of LoCoMo and LME. These depths are not a tuned advantage. Sweeping $k \in \{5, 10, 20, 40\}$ for the retrieval baselines (BM25, dense, Mem0) in the alone condition, accuracy on \textsc{StateMemBench} is flat across all depths (e.g.\ BM25 $20.0\%$ at every $k$; Mem0 $25$--$27\%$), so $k{=}10$ favors no method. On LongMemEval the retrieval baselines instead improve monotonically with $k$ (dense $35.0 \rightarrow 51.7\%$ and Mem0 $26.7 \rightarrow 45.0\%$ from $k{=}5$ to $40$; BM25 flat at $16.7\%$), so $k{=}20$ is, if anything, generous to the baselines rather than to \textsc{StateMem}.

\paragraph{Retrieval (BM25, \texttt{text-embedding-3-small}).}
Both ingest the conversation as per-turn chunks and call the LLM only at the answer step. BM25 is lexical retrieval via \texttt{rank\_bm25} (\texttt{BM25Okapi}); the dense
retriever uses cosine top-$k$ over OpenAI \texttt{text-embedding-3-small} embeddings.

\paragraph{Graph-RAG baselines (nano-graphrag, HippoRAG, LightRAG, GraphRAG).}
Four graph-structured retrieval systems that build an entity/relation graph over the conversation at ingest and retrieve over it at answer time. nano-graphrag and LightRAG build a local knowledge graph with community summaries; HippoRAG retrieves by personalized PageRank over an OpenIE graph (contriever embedder); GraphRAG (Microsoft) runs its CLI index/query pipeline per scenario. All use the panel's substrate model for extraction and answering, with OpenAI \texttt{text-embedding-3-small} for embeddings.

\paragraph{Mem0.}
Installed via PyPi package \texttt{mem0ai} with a ChromaDB vector store and sentence-transformers, using Mem0's default retriever. Mem0 uses a two-stage extraction and update phase, where the LLM can use tool calling during the update phase to keep the memories up to date \citep{mem0}.

\paragraph{A-Mem.}
Cloned from \texttt{agiresearch/A-mem}, default retriever. We relax its \texttt{json\_schema} response format to \texttt{json\_object} for DeepSeek compatibility. A-Mem enables dynamic, agent memory operations; it extracts structured notes, links them, updates linked memories, and retrieves relevant memories at test time \citep{xu2025amemagenticmemoryllm}.

\paragraph{LightMem.}
Cloned from (\texttt{zjunlp/LightMem}), default retriever. LightMem focuses on lightweight memory as the name would suggest and has three stages: compression, consolidation, and sleep-time update \citep{fang2025lightmem}.

\paragraph{MemoryOS.}
Installed via PyPi package \texttt{memoryos-pro} (installed \texttt{--no-deps} with \texttt{faiss-cpu}), default retriever. MemoryOS organizes memory into three tiers (long-, mid-, and short-term) and operates differently and dynamically on different kinds of memory \citep{kang-etal-2025-memory}.

\paragraph{A note on LightMem and MemoryOS scores.} Both systems score near floor on every benchmark we run, below any plausible level for published systems. We run the released implementations directly (official clones / PyPI, upstream-example configs; LightMem through its native vLLM provider with topic segmentation and a forced end-of-ingest flush), and roughly $60\%$ of their answers on \textsc{StateMemBench} are still non-answers---little relevant content surfaces at answer time under per-turn conversational ingest. Whether this reflects the methods' sensitivity to this workload or a residual integration gap, the rows are not evidence about best-case capability, so we report them for completeness and exclude them from comparative claims; no headline comparison depends on them.

\paragraph{Per-arm salience floor on Qwen.} Supporting \S\ref{sec:error}: gold counts on salience probes on the Qwen substrate, Set~A ($n{=}33$) / Set~B ($n{=}18$): long-context $5/33$ and $0/18$; Mem0 $1/33$ and $1/18$; nano-graphrag $5/33$ and $0/18$; GraphRAG $6/33$ and $0/18$; StateMem $0/32$ and $1/18$ (two StateMem Set~A probes were dropped at grading, hence $32$ rather than $33$). Every arm sits at or below $18\%$, versus $0.569$ for GPT-5.4-Nano long-context, so the salience collapse tracks the substrate, not the memory system. On DeepSeek, StateMem's salience is $0.216$ (Table~\ref{tab:state-mem-bench-full}).

\paragraph{Long-context models (Qwen-3.5-9B, Qwen-3.6-35B-A3B, GPT-5.4-Nano, DeepSeek-V4-Flash).}
The Qwen models are served locally via vLLM (Qwen-3.5-9B at 131072 token context; Qwen-3.6-35B-A3B reduced to 16384 after KV-cache OOM). Set~A scenarios (${\sim}3$k tokens) fit the reduced window comfortably, but the longest Set~B scenarios (${\sim}7$--$15$k tokens) approach or exceed it once the answer prompt is added and are tail-truncated, so the 35B long-context row should be read as truncation-affected on Set~B. (Its Set~A score, which no truncation touches, is also below the 9B model's, so truncation is not the sole cause of its weak showing.) GPT-5.4-Nano  is queried via the OpenAI API and DeepSeek-V4-Flash via the DeepSeek API; all model reasoning or thinking is disabled, matching the substrate configuration used by the memory baselines. All calls use temperature 0.
\begin{table*}[h]
\centering
\small
\caption{Closed-pool failure modes on DeepSeek-V4-Flash, $n{=}322$ (combined Set~A+B; errors are 0 for every method). Other-opt means an answer that was neither the drift target nor the gold truth but \textit{was} in the closed-pool plausible options (constructed at generation time). Off-pool means the method did not answer with a recognizable option ($\texttt{matched\_option}$ null). Drift-rate is the percentage of answers selecting the drift target over all answers (drift\,/\,$n$).}
\label{tab:drift_full_ds}
\begin{tabular}{lrrrrr}
\toprule
Method & Correct & \textbf{Drift} & Other-opt & Off-pool & Drift-rate \\
\midrule
Long-context (V4-Flash) & 48 & 206 & 4 & 64 & 64.0\% \\
Long-context (GPT5.4n)  & 89 & 159 & 1 & 73 & 49.4\% \\
\textbf{StateMem (ours)}  & \textbf{117} & 158 & 11 & 36 & 49.1\% \\
\abl{extraction-only}          & 56 & 226 & 6 & 34 & 70.2\% \\
\abl{+ supersession}           & 96 & 184 & 9 & 33 & 57.1\% \\
\abl{$-$ dependency-propagation} & 120 & 157 & 9 & 36 & 48.8\% \\
\abl{$-$ recompute-guidance}   & 97 & 179 & 10 & 36 & 55.6\% \\
Mem0  & 57 & 193 & 18 & 54 & 59.9\% \\
A-Mem & 64 & 181 & 5 & 72 & 56.2\% \\
LightMem & 4 & 48 & 0 & 270 & 14.9\% \\
MemoryOS & 8 & 83 & 1 & 230 & 25.8\% \\
nano-graphrag & 59 & 188 & 7 & 68 & 58.4\% \\
HippoRAG & 59 & 175 & 12 & 76 & 54.3\% \\
LightRAG & 30 & 150 & 5 & 137 & 46.6\% \\
GraphRAG (MS) & 56 & 158 & 8 & 100 & 49.1\% \\
BM25 & 46 & 193 & 5 & 78 & 59.9\% \\
Dense & 66 & 177 & 13 & 66 & 55.0\% \\
No memory & 1 & 18 & 0 & 303 & 5.6\% \\
\bottomrule
\end{tabular}
\end{table*}

\begin{table*}[h]
\centering
\small

\caption{Closed-pool failure modes on Qwen-3.5-9B, $n{=}322$ (combined Set~A+B; two cells $n{=}320$ after 2 grading drops, errors otherwise 0). Other-opt means an answer that was neither the drift target nor the gold truth but \textit{was} in the closed-pool plausible options (constructed at generation time). Off-pool means the method did not answer with a recognizable option ($\texttt{matched\_option}$ null). Drift-rate is the percentage of answers selecting the drift target over all answers (drift\,/\,$n$).}
\label{tab:drift_full_qwen}
\begin{tabular}{lrrrrr}
\toprule
Method & Correct & Drift & Other-opt & Off-pool & Drift-rate \\
\midrule
Long-context-9B  & 48 & 211 & 4 & 59 & 65.5\% \\
Long-context-35B & 31 & 215 & 6 & 68 & 67.2\% \\
\textbf{StateMem (ours)} & \textbf{75} & 207 & 8 & 32 & 64.3\% \\
\abl{extraction-only}          & 50 & 239 & 9 & 24 & 74.2\% \\
\abl{+ supersession}           & 55 & 230 & 10 & 27 & 71.4\% \\
\abl{$-$ dependency-propagation} & 71 & 212 & 13 & 24 & 66.2\% \\
\abl{$-$ recompute-guidance}   & 67 & 220 & 7 & 28 & 68.3\% \\
Mem0     & 48 & 202 & 4 & 68 & 62.7\% \\
A-Mem    & 41 & 206 & 4 & 71 & 64.0\% \\
LightMem & 6  & 103 & 0 & 213 & 32.0\% \\
MemoryOS & 8  & 112 & 1 & 201 & 34.8\% \\
nano-graphrag & 46 & 191 & 7 & 78 & 59.3\% \\
HippoRAG & 42 & 194 & 11 & 75 & 60.2\% \\
LightRAG & 37 & 151 & 20 & 114 & 46.9\% \\
GraphRAG (MS) & 72 & 185 & 4 & 61 & 57.5\% \\
BM25 & 40 & 196 & 7 & 79 & 60.9\% \\
Dense & 42 & 209 & 17 & 54 & 64.9\% \\
No memory & 2 & 78 & 0 & 242 & 24.2\% \\
\bottomrule
\end{tabular}
\end{table*}
\section{Limitations}
\textsc{StateMemBench} uses synthetically-generated data, which limits its ability to genuinely reflect real world use cases for agent memory systems; its traps also come from the same lazy-reader policy family that \textsc{StateMem} is built to counter, so margins on our own benchmark are best read as an upper bound, with the external benchmarks as the generalization check. Additionally, while we show that drift happens even before models and memory systems approach the ``true'' long horizon, our scenarios (${\sim}3$k tokens in Set~A, ${\sim}7$--$15$k in Set~B) are much shorter than the typical modern LLM context window, which limits their ability to truly stress test long horizon state drift.

On the evaluation side, the grading judge shares a model family with one substrate and scenarios are rendered by Claude-family models. Additionally, for cost reasons, both substrates are small-to-mid-scale models, and behavior under frontier answerers is untested. Reported numbers are single runs. Also, the drift-prevalence rates of \S\ref{sec:prevalence} are LLM-judge labels that human annotators tend to undercut. \textsc{StateMem} itself extracts state units at every turn (${\sim}165$--$600$ encoder calls per scenario), so it uses significantly more resources than a long-context LLM on its own. While none of our released data is particularly sensitive, increasing LLM agent capabilities is always worth some consideration. Better state tracking might encourage users to become more reliant on LLM agents, which could have negative effects on users, especially in relatively serious domains like personal finance.

\section{AI Usage Disclosure}
We designed all experiments and analyses ourselves. We used AI coding agents (e.g., Claude Code) to assist in implementing experiment code and to help manage compute infrastructure (e.g., provisioning and running AWS instances). We also used LLMs for polishing writing, suggesting organizational structure for the paper, and generating LaTeX table code. All scientific claims, experimental designs, and conclusions are our own, and the authors take full responsibility for the content of the paper.
\end{document}